%% file: acl_latex.tex
\documentclass[11pt]{article}

\usepackage[preprint]{acl}

\usepackage{times}
\usepackage{latexsym}
\usepackage[T1]{fontenc}
\usepackage[utf8]{inputenc}
\usepackage{microtype}
\usepackage{inconsolata}
\usepackage{graphicx}

\usepackage{xcolor}
\usepackage{natbib}
\usepackage{booktabs}
\usepackage{multirow}
\usepackage{tabularx}
\usepackage{caption}
\usepackage{fancyvrb}
\usepackage{fvextra}
\newcommand{\beforefinetuning}{No Profile}
\newcommand{\syntheticbaseline}{Synthetic}
\newcommand{\persona}{Open-Ended}
\newcommand{\noprofile}{No Profile}
\usepackage{subcaption}
\usepackage[most]{tcolorbox}

\newtcolorbox{promptbox}[1][]{
  colback=red!5,
  colframe=red!60,
  boxrule=0.5pt,
  arc=2pt,
  left=6pt,
  right=6pt,
  top=6pt,
  bottom=6pt,
  title=#1,
  fonttitle=\bfseries,
  breakable
}

\title{Behaviorally Grounded User Profiles from the Wild for\\Personalized Alignment and Multi-Perspective Reasoning}

\author{Yuxuan Li\thanks{Corresponding author: \texttt{yuxuan.li1@uwaterloo.ca}.} \\
  University of Waterloo \\\And
  Victor Zhong \\
  University of Waterloo \\
  \\\And
  Ehsan Kamalloo \\
  ServiceNow AI Research \\
  }

\begin{document}
\maketitle

\input{body}

\bibliography{custom}

\appendix
\input{appendix}

\end{document}

%% file: body.tex
\begin{abstract}
Persona-driven techniques increasingly adapt large language models (LLMs) to diverse contexts. However, existing methods predominantly rely on rigid, synthetic personas that flatten individual variation, rely on stereotypes, and miss the nuanced signals driving actual human preferences.
We introduce \textit{profile behavioral grounding}, a framework for extracting open-ended, high-fidelity user profiles directly from authentic, anonymized social media posts.
We evaluate these profiles across two paradigms: train-time personalization via supervised finetuning (SFT) and non-parametric test-time multi-perspective reasoning.
Across complex recommendation and open-ended query benchmarks, behaviorally grounded profiles consistently improve base models and outperform synthetic profile baselines, driving stronger parametric alignment and enabling richer, multifaceted reasoning.
Our findings establish open-ended, behavior-derived profiles as a highly diverse and effective foundation for the next generation of personalized language systems. Our code base is available at \url{https://github.com/ServiceNow/behavior-grounding}.
\end{abstract}

\begin{figure*}
    \centering
    \includegraphics[width=1\linewidth]{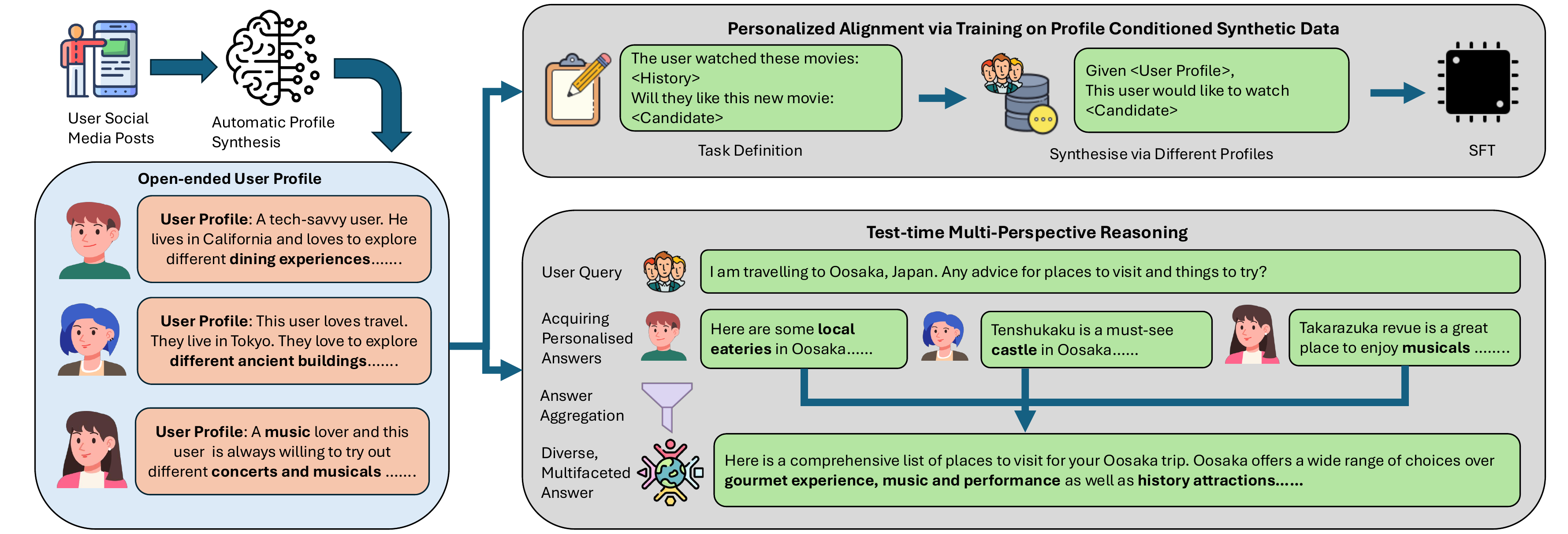}
    \caption{Our framework for user profile synthesis and utilization. We extract open-ended user profiles from user behaviour sequences (e.g. social media posts). With these diverse, realistic profiles, we synthesize SFT data to parametrically align models for personalization tasks. We additionally study test-time multi-perspective reasoning to elicit diverse, multi-faceted responses without weight updates.}
    \label{fig:figure_1}
\end{figure*}

\section{Introduction}
As large language models (LLMs) are increasingly deployed as personal assistants, conversational agents, and recommendation engines, their ability to adapt to distinct user contexts and preferences has become a critical frontier \citep{salemi-etal-2024-lamp,zhao2025llms}.
To achieve this personalization, recent systems heavily rely on persona-driven techniques \citep{personalizedjudge2024, personalizedllm}.
These user representations are now foundational across a broad spectrum of tasks, ranging from preference prediction and recommendation \citep{salemi-etal-2024-lamp} to aligning open-ended generation with diverse human values \citep{kirk2024prism}.
However, the efficacy of this personalization is fundamentally bottlenecked by the fidelity of the user representation itself.

Most prior work models users via synthetic profiles constructed from a rigid, small set of categorical attributes, such as age, gender or nationality \cite{hao2025evaluating, persona_viability}.
While computationally convenient and easy to condition on, these fully synthetic personas flatten complex individual variations into predefined buckets.
In practice, they often rely on stereotypical correlations~\citep{cheng2023compost} and fail to capture the nuanced, idiosyncratic behavioral signals that drive actual human preferences.
In the real world, a user's intent is rarely neatly organized into an attribute table; rather, it must be inferred from distal, unstructured, and messy traces of authentic behavior.

To bridge this gap, we introduce \textit{behavioral grounding}: a framework for extracting open-ended, highly realistic user profiles directly from authentic social media activity.
Rather than relying on LLM-hallucinated stereotypes, our pipeline synthesizes short, coherent textual bios from observed behavioral sequences (e.g., historical user posts), effectively cleaning and summarizing noisy social signals into rich intermediate representations.
Our contribution is therefore at the level of user representation rather than a new optimizer, adapter, or aggregation algorithm: by keeping the downstream adaptation mechanisms deliberately simple, we isolate whether the source of the profile itself---authentic behavioral traces rather than synthetic persona generation---changes personalization quality.
We argue that these behavior-grounded profiles offer a superior and more realistic foundation for personalization compared to synthetic categorical personas.

We evaluate the utility of these open-ended profiles across two distinct operational paradigms.
First, we explore \textit{train-time personalization}, utilizing profile-conditioned synthetic data to perform supervised finetuning (SFT) for downstream tasks, including complex recommendation and open-ended question answering.
Second, we introduce \textit{test-time multi-perspective reasoning}, a non-parametric approach where multiple semantically relevant user profiles are sampled as candidate perspectives, aggregating their unique viewpoints into a comprehensive, multi-faceted final response without requiring weight updates.

Across recommendation domains, open-ended query benchmarks, and base models, our results demonstrate that behavioral grounding significantly enhances downstream performance in both paradigms.
Our analysis reveals that these gains are driven by the broader diversity and lower concentration of our extracted profiles compared to synthetic profile baselines, allowing models to generate richer and more nuanced outputs.
Our main contributions are summarized as follows:

\begin{itemize}
\item We introduce a behavioral grounding framework designed to extract and distill open-ended, human-readable user profiles from unstructured, authentic social media activity.
\item We demonstrate the versatility of these profiles across two distinct paradigms: train-time personalization via SFT data synthesis, and test-time multi-perspective reasoning for multifaceted answers.
\item We provide empirical and analytical evidence showing that behavior-grounded profiles offer a more diverse, representative, and effective alternative to synthetic personas in personalized language systems.
\end{itemize}

\section{Related Work}

\paragraph{Persona-Driven Augmentation and Alignment}
Recent work uses persona-conditioned synthesis for data augmentation, scaling to large LLM-generated persona banks or targeting cultural and regional diversity \citep{persona_hub,culturellm,culturepark}.
Benchmarks such as CulturalBench, BLEnD, and CASA stress-test models on regional commonsense \citep{culturalbench,blend,casa}, but HCI and alignment studies caution that synthetic personas can introduce artifacts, stereotypes, and user misrepresentation \citep{persona_viability,kirk2024benefits}.
Our work instead emphasizes \emph{behavioral grounding}: extracting profiles from authentic social media activity to connect broad profile diversity with real-world human fidelity.
\paragraph{LLM Adaptation to User Contexts}
Personalization research adapts LLM behavior using various signals, including explicitly stated attributes, customizable interfaces \citep{clo_chat}, conversational histories \citep{one_chatbot_per_person}, and high-intent interaction histories for recommendation \citep{llmrec, mgshopdial}.
Existing benchmarks and alignment studies measure profile sensitivity, social character fidelity, and preference-aware generation \citep{wu2024socialsignals,personalllm,patching2025,prism,urs2024,personalizedjudge2024}.
Recent systems summarize, retrieve, or internalize task-specific histories, including summary-augmented retrieval for LaMP-style tasks \citep{richardson2023integrating}, PRIME's dual-memory reasoning \citep{zhang2025prime}, per-user PEFT in OPPU \citep{tan2024democratizing}, parameterized memory injection in MiLP \citep{zhang2024personalizedmilp}, and TSUBASA's evolving memory with context distillation \citep{zhang2026tsubasa}.
These works primarily optimize adaptation to in-domain histories for a known user; we instead test whether distal social traces can be distilled into reusable open-ended profiles, holding adaptation fixed to isolate the effect of profile source.

\begin{figure*}[!th]
    \centering
    \includegraphics[width=1\linewidth,trim={80pt 15pt 150pt 0pt},clip]{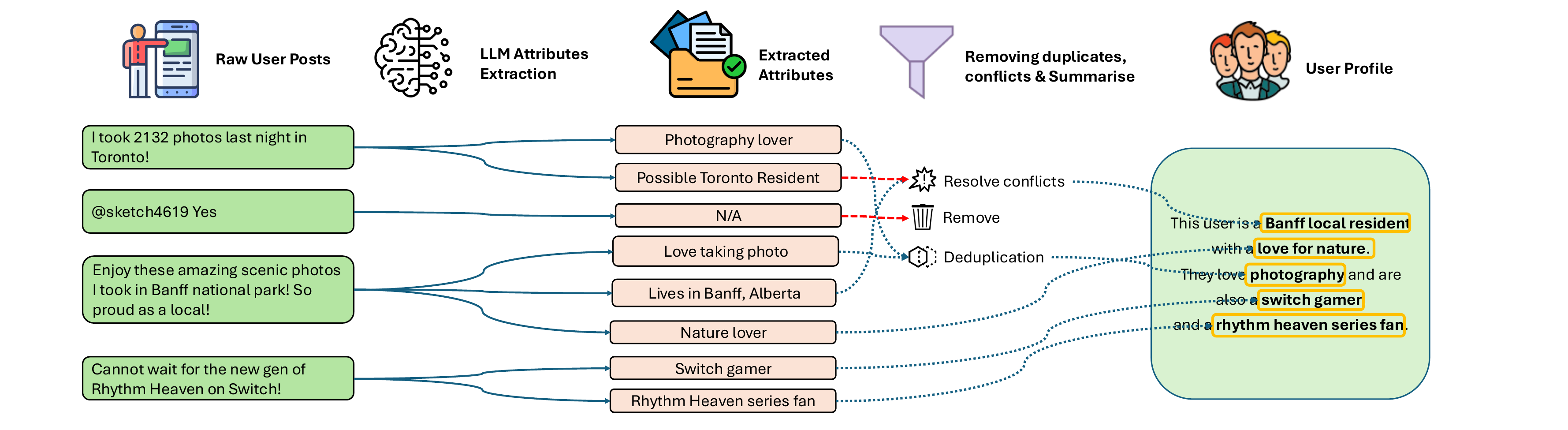}
    \caption{Our pipeline to synthesize open-ended, behaviourally grounded user profiles from wild social media data. We iterate over user posts and extract possible user attributes. We then remove duplicates and contradictions and summarize user attributes as a coherent user profile.}
    \label{fig:persona_extraction}
\end{figure*}

\paragraph{Test-Time Scaling and Diverse Perspectives}
Beyond parametric updates, multi-persona prompting \citep{wang2024selfprompting} and Mixture-of-Agents frameworks use role-play or self-collaboration to elicit stronger reasoning and broader solution spaces.
Their diversity typically comes from arbitrary roles, model variation, or temperature scaling; we instead use behaviorally grounded profiles as distinct ``experts'' for \emph{perspective sampling}.
This scales test-time compute by crowdsourcing multifaceted perspectives for complex, open-ended queries.

\section{Methodology: The Open-Ended Profiling Framework}
As shown in Figure~\ref{fig:figure_1}, our approach extracts rich behavioral signals from social media (Section~\ref{sec:extraction}) to unlock two distinct capabilities: parametric adaptation for personalized alignment (Section~\ref{sec:parametric}), and non-parametric test-time multi-perspective reasoning (Section~\ref{sec:nonparametric}).

\subsection{Behaviorally Grounded Profile Extraction}
\label{sec:extraction}

We propose an automatic pipeline to extract open-ended user profiles from unstructured user activities, specifically social media posts.
While processing large-scale social data incurs a higher initial computational cost than generating cheap synthetic profiles, we argue this one-time extraction yields a reusable, high-fidelity profile bank that significantly improves downstream task performance.
In this study, we utilize a Bluesky dataset \cite{dale_2m_bluesky_2024} consisting of 2 million posts collected from 2023 to 2024. 
Figure~\ref{fig:persona_extraction} illustrates the extraction pipeline.
First, we process each user post with an LLM to extract candidate attributes and preferences expressed in the text.
For example, if a user posts about video games, we extract \emph{video game fan} as a latent attribute. After this stage, each user is associated with a collection of short textual descriptors.
Second, we clean the descriptor set by removing redundancy and resolving inconsistencies. 
Concretely, through prompting and lexical filtering, we de-duplicate near-identical statements (e.g. user posts multiple times about video games) and filter descriptors that are contradictory or unsupported by the overall post history (e.g. mutually incompatible current-location descriptors without temporal evidence).
Third, we use an LLM to summarize the filtered descriptions into a short, cohesive bio that serves as the final user profile.\footnote{In practice, steps two and three are completed within a single LLM call.}
To ensure user privacy and prevent mosaic-effect re-identification, this summarization step explicitly abstracts granular spatiotemporal details into broader behavioral patterns (detailed further in our Ethical Considerations section).

\subsection{Personalized Alignment via SFT}
\label{sec:parametric}
To adapt a model to ground its output on our open-ended profiles without relying on costly human-annotated preference data, we introduce a profile-conditioned data synthesis pipeline for Supervised Finetuning (SFT).
Formally, let $U$ denote the extracted open-ended user profile, $H$ the user's historical interaction sequence, and $I$ a candidate input item.
We prompt a strong teacher LLM to estimate the preference distribution $P(Y \mid U, H, I)$. In practice, we decode the most probable sequence $Y$ using greedy decoding to represent the target response (e.g., a binary like/dislike label or an open-ended answer). We then serialize these deterministic targets into standard next-token prediction instruction-tuning pairs (e.g., \texttt{Given <$H, I$> or <$U, I$>, predict <$Y$>}).
To this end, we convert each downstream task into a personalized task by incorporating a user profile into the instructions. We construct synthetic training data by instructing a strong teacher LLM to predict the user's likely response or preference given a user profile.
All prompt templates and synthesis steps for each of our evaluation domains (recommendation, conversational preference, and open-ended QA) are detailed in Appendix~\ref{sec:prompt}.

\subsection{Test-time Multi-Perspective Reasoning}
\label{sec:nonparametric}
While our SFT experiments focus on aligning a model with a \emph{specific} user profile, our test-time multi-perspective reasoning leverages the \emph{diversity} of our profile bank to improve general response quality without any parametric updates.
This setting is not target-user personalization: the goal is to answer an open-ended query with broad coverage across plausible viewpoints, not to predict the preferences of the person asking the question.
We therefore do not condition on the inquirer's own profile during evaluation.

We employ a \textit{Mixture of Perspectives} approach: given a generic or open-ended user query (e.g., ``\emph{Tips for traveling to Osaka}''), 
we sample an initial set of user profiles from our dataset and use an LLM to discard non-relevant ones (e.g., it is highly unlikely that a German resident would inquire about Costco pricing in dollars, as Costco does not operate there), retaining $k$ profiles with potentially relevant interests.
The base model generates a candidate response conditioned on each of these distinct personas, yielding a spectrum of complementary viewpoints. Finally, we employ another stronger LLM that operates as an aggregator to synthesize these diverse perspectives into a single, comprehensive answer. 
Specifically, the aggregator is instructed to identify contradictions, attribute conflicting advice to differing user contexts, and weave the distinct viewpoints into a unified, multifaceted reply without explicitly exposing the underlying personas to the user.
Our goal here is to examine whether our open-ended profiles can serve as distinct ``experts'' to reduce hallucination, increase diversity, and improve coverage for complex queries at inference time.
The prompt examples are shown in Appendix~\ref{sec:prompt}.

\section{Experiments and Analysis}

Our experimental design aims to answer three core research questions:
\begin{itemize}
    \item \textbf{RQ1 (Efficacy):} Does finetuning with open-ended, behavior-derived user profiles improve model alignment and personalization on downstream tasks?
    \vspace{-0.1in}
    \item \textbf{RQ2 (The Grounding Hypothesis):} Do profiles grounded in actual user behavior outperform task-relevant synthetic profiles? 
    \vspace{-0.1in}
    \item \textbf{RQ3 (Perspective Scaling):} Can test-time multi-perspective reasoning across diverse user profiles improve performance on complex, open-ended queries?
    
\end{itemize}

\subsection{Evaluation Benchmarks}
We evaluate model performance on downstream tasks that require diverse perspectives and personalization.
Importantly, all evaluation targets are grounded in real user data and human annotations.

\paragraph{RecBench \cite{liu2025can}}
is a benchmark for LLMs on recommendation tasks.
It consists of a diverse set of recommendation scenarios spanning multiple domains, such as movies, music, and books.
In this work, we focus on few-shot, pair-wise recommendation tasks: given a user interaction history, can the LLM predict if the user will like a new candidate item? 
We report the F1 score of the prediction as our core metric. In our evaluation, we used a subset of 10K for our evaluation.

\paragraph{URS Bench \cite{urs2024}}
is a multi-intent user query dataset annotated by human participants that covers six types of user intents.
We consider the three query types most tied to personalization: Ask for Advice, Seek Creativity, and Leisure and we only consider English queries.
We follow URS to conduct evaluation using an LLM-as-a-judge over criteria such as user satisfaction and factuality.
Unlike the original URS protocol, we additionally provide the judge with the user profile to assess personalization quality.
We report the average 1-to-10 score from the LLM judge.
To assess the reliability of this metric, we conduct a human study, finding that LLM judge scores align closely with human ratings (Appendix~\ref{sec:human_judge_validation}).

\subsection{Baselines}
\label{sec:baselines}
To evaluate the value of our profiling framework, we compare against two primary baselines:

\paragraph{{\beforefinetuning}:} The model is prompted to complete the task using only the context and the prompt, without any open-ended user profile provided and no additional finetuning.

\paragraph{Task-Relevant Synthetic Profiles Baseline:} To isolate the value of \emph{behavioral grounding}, we construct a baseline of purely synthetic user profiles. 
Unlike our open-ended profiles derived from real activity, these baselines are generated by prompting an LLM to sample with different seeds and a temperature of 0.7 to obtain plausible user personas relevant to the specific downstream task (e.g., ``\textit{Generate an open-ended profile for a user given examples}'').
While recent works~\citep{persona_hub,persona_viability} have proposed highly complex, multi-agent pipelines or massive taxonomic trees to force synthetic diversity, we intentionally utilize standard, out-of-the-box persona generation practices (varying seeds and temperature).
This design choice is crucial for ablation: it isolates the specific value of authentic behavioral grounding. 
Introducing heavily engineered synthetic pipelines would introduce confounding variables, making it difficult to determine whether downstream gains stem from realistic behavioral nuances or merely from complex prompt engineering artifacts.
Prompt templates for synthetic baseline generation are in Appendix~\ref{sec:prompt}.
To confirm that the resulting diversity gap is inherent to synthetic generation rather than an artifact of our procedure, in Appendix~\ref{sec:stronger_synthetic_ablation}, we evaluate several stronger persona generation methods, along with an ablation study over the stages of our own pipeline.

\subsection{Implementation Details}
We conduct experiments with 5 different models across three families: Qwen3-8B, Qwen3-14B, Qwen3-32B, Gemma3-4B, and Olmo3-7B-Instruct.
For Qwen3 series models, we turn off the extended thinking capabilities during evaluation and finetuning.
This ensures equitable compute comparisons with standard models and isolates the performance gains strictly to the profile conditioning, rather than the model's internal reasoning loops.
The user profile extraction and SFT data synthesis are performed using GPT-OSS-120B.
We also utilize GPT-OSS-120B as our LLM judge for URS Bench evaluation.
Complete hyperparameters and training configurations are provided in Appendix~\ref{sec:experimental_details}.

\subsection{Results: Personalized Alignment via SFT}
\input{tables/sft_table}
The personalized alignment via SFT results are presented in Table~\ref{tab:task_comparison}.
For recommendation tasks, \persona{} improves over \syntheticbaseline{} in 13 of 15 RecBench cells and is best on URS average for all five models.
The gains are not only caused by weak base behavior in Olmo3 7B: for example, Gemma3 4B improves from 0.309 to 0.643 F1 on Books, Qwen3 14B improves from 0.308 to 0.632, and Qwen3 32B improves from 0.569 to 0.658.
Olmo3 7B shows the largest absolute change because the unfinetuned model tends to answer ``No'' for most inputs, but the broader pattern also holds for stronger non-degenerate base models.

In the URS benchmark, we observe that finetuned models achieve the highest performance when using our open-ended user profiles, illustrating the benefits of realistic user representations.
For example, in Gemma3 4B, \syntheticbaseline{} results in a small performance drop while \persona{} results in a score increase of 0.82.
We also find that for the intent of asking for advice, \persona{} almost always has the highest performance, this correlates with the fact that asking for advice often requires more personalization than the other two intents.

While \persona{} generally yields strong gains, we observe neutral or negative results in two boundary cases (PRISM Preference Prediction and RecBench Lastfm), discussed in Appendix~\ref{sec:additional_prism_lastfm}.
In PRISM, performance plateaued because compressing open-ended bios into PRISM's required categorical format caused severe information loss.
In Lastfm, finetuning hurt performance due to profile domain sparsity (e.g., lacking music-related keywords) and substantial downstream distribution shift.
These cases highlight that open-ended profiling is constrained by forced categorical formatting and highly niche domains.

We also examine whether our SFT alignment impacts the general capabilities of the finetuned models and find that these capabilities are largely preserved, as shown in Table~\ref{tab:general-capabilities} (Appendix~\ref{app:general-capabilities}).

\subsection{Results: Test-Time Multi-Perspective Reasoning}
We also run a Test-Time Multi-Perspective Reasoning experiment, where, instead of performing parametric updates, we sample multiple answers conditioned on different user profiles and synthesize them into multifaceted responses.
We similarly evaluate this experiment on URS Bench. In these experiments, due to the limited context window of the summarizing model, we sample 5 answers for each query.
For \persona{}, we sample answers from ``related user profiles,'' conducting an additional round of annotation to ensure the user could plausibly pose such a query.
For \syntheticbaseline{}, a similar process cannot be done as the purely synthetic baseline shows lack of diversity and consists largely of homogeneous profiles.
\input{tables/test_time_scaling}
The results are shown in Table~\ref{tab:urs_only}. \persona{} generally outperforms \noprofile{} and \syntheticbaseline{}. For example, for Olmo3 7B, \persona{} yields up to +0.50 score improvement, and for Qwen3 32B it yields +0.52.
This trend is consistent across different families of models or sizes of models, confirming that perspective sampling is a robust, model-agnostic technique for acquiring multifaceted answers.  
We also observe that \syntheticbaseline{} improves over \noprofile{}, but by less than \persona{}, consistent with synthetic profiles being more homogeneous.
Overall, to answer \textbf{RQ3}, our results demonstrate that test-time perspective sampling across realistic and diverse user profiles improves performance on open-ended user queries.
Appendix~\ref{sec:temperature_ablation} further shows that increasing decoding temperature alone does not substitute for grounded perspectives.

\paragraph{Is Sampling from Relevant Profiles Necessary?}
In Test-Time Multi-Perspective Reasoning, we generate diverse perspectives by grounding candidate answers in \emph{Relevant} user profiles.
To isolate the effect of sampling from related profiles, we replace them with randomly sampled ones and generate answers accordingly. Results are shown as ``Random'' in Table~\ref{tab:urs_only}.
We find that ``Random Profiles'' consistently outperforms the synthetic baseline across models, 
confirming that diversity in user perspectives alone is helpful.
However, it is consistently lower than Open-Ended Profile, where only related user profiles are sampled, with Qwen3 8B being the only exception, due to the randomness in user profile selection in Random Profile baseline.

\paragraph{Is high-temperature sampling enough to replace profile-grounded perspectives?}
To answer this question, Appendix~\ref{sec:temperature_ablation} compares profile-grounded multi-perspective reasoning against a high-temperature sampling baseline without profile conditioning.
The results show that high-temperature sampling performs worse, thereby
suggesting that the gains come from grounded behavioral perspectives rather than output diversity alone.

\subsection{Analysis: Behavioral Grounding vs. Synthetic Profiles}
\label{sec:analysis}
\begin{figure}[th]
    \centering
    \includegraphics[width=1\linewidth]{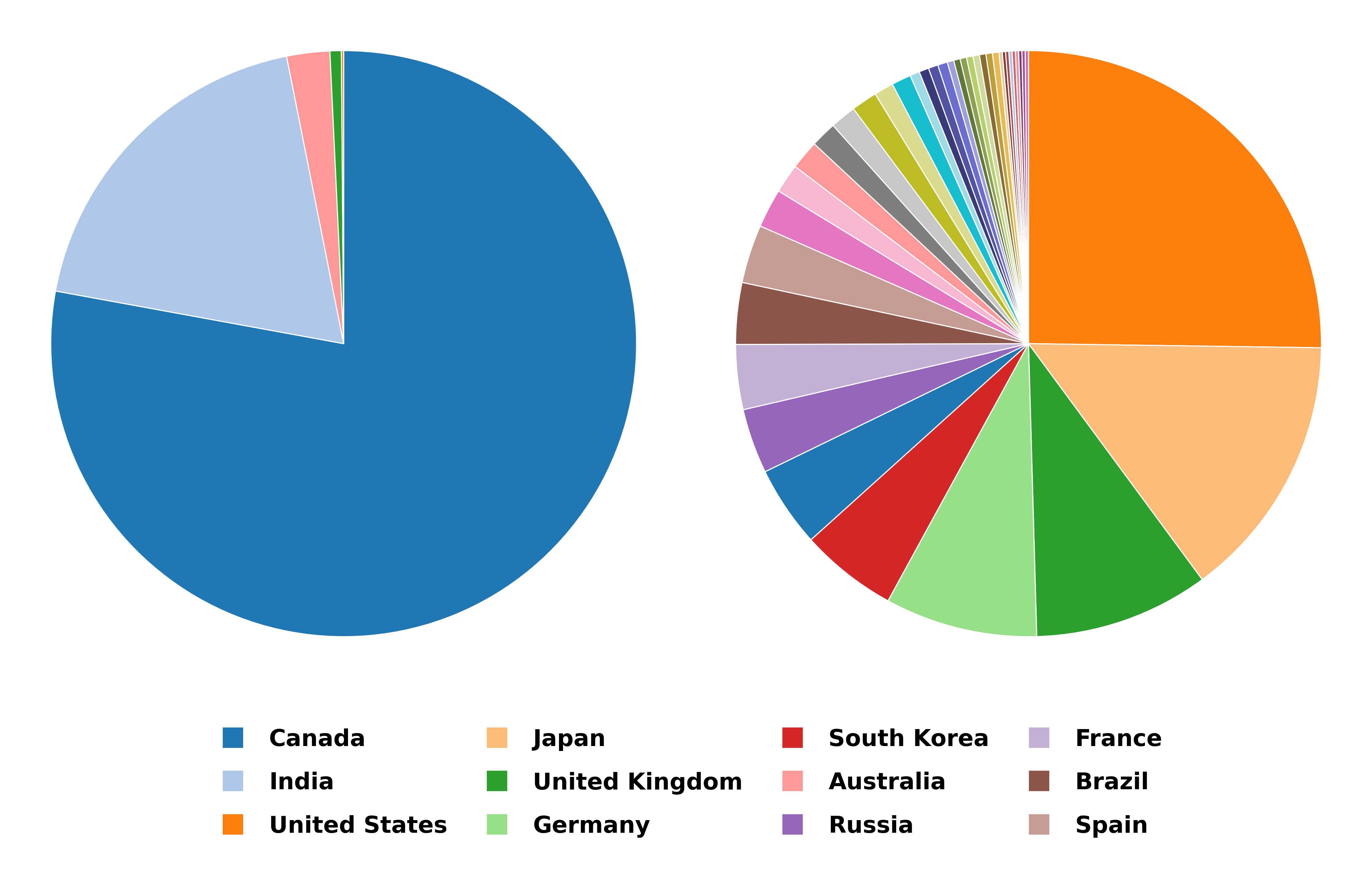}
    \vspace{-0.3in}
    \caption{Birth location distribution of Baseline Synthetic User Profiles (Left) and Our Open-ended (Right). We only show the legends of the top 12 countries here due to limited space.}
    \label{fig:persona_location}
    \label{fig:baseline_location}
    
\end{figure}

\paragraph{User Profile Distribution Analysis:}
Following \citet{kirk2024prism}, we extract categorical attributes from both profile sets to analyze their distributions. As shown in Figures~\ref{fig:baseline_location}, synthetic profiles exhibit severe demographic collapse, skewing heavily toward a narrow subset of nationalities and demographics. In contrast, our open-ended profiles maintain a highly representative long-tail distribution across all extracted attributes. This diversity is quantified in Table~\ref{tab:categorical-entropy-comparison}, where the categorical entropy of our behaviourally grounded profiles significantly exceeds the synthetic baseline, closely rivalling human-annotated survey data from PRISM. We present additional results on representativeness and faithfulness of our user profiles with respect to the underlying behavioral data in Appendix~\ref{app:representative} and Appendix~\ref{app:faithful}, respectively.

\input{tables/profile_categorical_entropy}

We also conduct a PCA analysis with our two sets of user profiles, the embeddings from Qwen3 8B embedding, shown in Appendix~\ref{sec:pca_plots}.
As our findings suggest, our user profiles are significantly more diverse and representative in their distribution than purely synthetic user profiles.

\input{tables/mauve_score}
\paragraph{How Close Is Data from Open-Ended Profiles to Downstream Ground-Truth?}
To quantify how well our synthesized SFT data, generated from open-ended profiles, approximates downstream data, we measure the MAUVE score~\citep{mauvescore}, a divergence-based metric that compares the distribution of model-generated text to a reference corpus using text embeddings.
Specifically, we embed both the synthesized SFT data and the ground-truth downstream data with Qwen3 8B, then compute MAUVE between two pairs: (1) SFT data generated from open-ended profiles vs. ground-truth, and (2) SFT data generated from synthetic profiles vs. ground-truth. 
Because URS queries are open-ended and lack ground-truth references, this analysis is done only on RecBench and PRISM (additional results for PRISM preference prediction are shown in Appendix~\ref{sec:additional_prism_lastfm}).
Table~\ref{tab:mauve_input_only} reports MAUVE scores averaged across 6 seeds. 
The scores indicate that both synthetic variants remain far from ground-truth, consistent with prior findings on the limitations of LLM-generated corpora~\citep{ren2025fewshotllmsyntheticdata}.
However, open-ended profiles yield consistently higher MAUVE scores than synthetic profiles across all domains, suggesting that grounding generation in real user language narrows the distributional gap.
For instance, in RecBench Pens (news recommendation), MAUVE increases from 0.013 to 0.090, a nearly $7\times$ improvement over the synthetic baseline.
This aligns with our SFT findings, where \persona{} consistently outperforms \syntheticbaseline{} (additional results on PRISM are reported in Appendix~\ref{sec:additional_prism_lastfm}).

Furthermore, we visualize the embedding space of synthesized SFT data using PCA (Appendix~\ref{sec:pca_plots}). The plots show that data generated from open-ended profiles often spans a broader region of the embedding space than the synthetic baseline, suggesting better coverage of the ground-truth distribution.

\subsection{Error Analysis}
To better understand the limitations of our framework, we categorize the most frequent failure modes observed across our two evaluation paradigms: parametric alignment (on RecBench/URS Bench) and non-parametric test-time scaling (on URS Bench). 

\subsubsection{RecBench SFT Failures}
\paragraph{Serendipitous and Multi-Interest Behavior:}
The finetuned models can over-rely on dominant category preferences, helping in-distribution cases but hurting serendipitous interactions~\citep{serendipity} and users with competing interests~\citep{li2019mind,cen2020controllable}.
For example, a dedicated sports reader may still click financial news, but such cross-category behavior is underrepresented in the synthesized training data.

\paragraph{Spurious Attributes:}
Models sometimes treat incidental item attributes as preference signals; e.g., a user who watched several Netflix movies from a narrow year range may receive recommendations confined to that period, even though release year is unlikely to define the user's taste.

\subsubsection{URS SFT Errors}
\paragraph{Knowledge Limits:}
Personalization improves fluency and perceived relevance, but it does not add external grounding for recent events, specialized domains, or location-specific recommendations; base and finetuned models therefore still produce generic or hallucinated answers to such queries.

\paragraph{Over Personalization:}
The models sometimes treat profile cues as obligatory even when the query does not require them.
For example, one hazardous-incidents query was framed through the user's interests in \textit{Barbie} and \textit{Tamako Market}, even though correctness and grounding should dominate.

Overall, finetuning often makes responses more fluent and tailored, but can induce unnecessary profile-based framing in direct factual queries.

\subsubsection{Multi-Perspective Reasoning Failures}

\paragraph{Language Drift and Persona Leakage:}
When English queries are answered through non-English-speaking profiles, synthesized answers can drift into another language.
The aggregator also sometimes retains irrelevant profile-specific details, cluttering broadly applicable answers with unhelpful personalization.

\paragraph{Safety Over-Propagation:}
Sampling multiple perspectives increases the chance that one candidate triggers a safety refusal, which the aggregator may propagate into the final answer.
These failures show that aggregation must preserve useful diversity while filtering language drift, irrelevant persona details, and overly conservative refusals.

\section{Conclusion}
We introduced a behavioral grounding framework that extracts open-ended user profiles from authentic social media activity rather than synthetic personas.
Across train-time SFT and test-time multi-perspective reasoning, these profiles improve personalization and exhibit structural diversity closer to human-annotated survey data.
Despite remaining challenges such as persona leakage, over-personalization, and domain sparsity, our results support behavior-grounded profiles as a high-fidelity intermediate representation for personalized language systems.

\input{limitations}

%% file: tables/sft_table.tex
\begin{table*}[!ht]
\centering
\caption{Performance comparison across base models, models finetuned with task-relevant \syntheticbaseline profiles, and models finetuned with our open-ended behavioral profiles. For URS, Creativity, Advice and Leisure corresponds to Seek Creativity, Ask for Advice and Leisure in URS Bench tasks. For RecBench, we report F1 score and for URS, we report 1 to 10 LLM judge score. Highest performing scores are highlighted in bold font.}
\label{tab:task_comparison}
\small 
\setlength{\tabcolsep}{4pt}
\begin{tabularx}{\textwidth}{l c XXX XXXX}
\toprule
\multirow{2}{*}{\textbf{Model}} & \multirow{2}{*}{\textbf{Variant}} & \multicolumn{3}{c}{\textbf{RecBench}} & \multicolumn{4}{c}{\textbf{URS}} \\
\cmidrule(lr){3-5} \cmidrule(lr){6-9}
& & Netflix & Books & News & Leisure & Creativity & Advice & \textbf{Avg.} \\
\midrule

\multirow{3}{*}{Qwen3 8B} 
& \beforefinetuning & 0.421 & 0.515 & 0.318 & 5.48 & 5.31 & 5.99 & 5.59 \\
& \syntheticbaseline & 0.420 & 0.625 & 0.319 & 6.34 & 6.72 & 7.08 & 6.72 \\
& \persona & \textbf{0.450} & \textbf{0.649} & \textbf{0.322} & \textbf{6.76} & \textbf{7.40} & \textbf{7.65} & \textbf{7.27} \\
\addlinespace 

\multirow{3}{*}{Qwen3 14B} 
& \beforefinetuning & 0.419 & 0.308 & 0.303 & \textbf{7.49} & 7.90 & 8.06 & 7.82 \\
& \syntheticbaseline & 0.416 & 0.538 & \textbf{0.327} & 7.06 & 7.54 & 7.91 & 7.50  \\
& \persona & \textbf{0.459} & \textbf{0.632} & 0.321 & 7.29 & \textbf{8.10} & \textbf{8.27} & \textbf{7.88} \\
\addlinespace

\multirow{3}{*}{Qwen3 32B} 
& \beforefinetuning & 0.403 & 0.569 & 0.308 & 6.79 & 7.09 & 6.90 & 6.93 \\
& \syntheticbaseline & 0.427 & 0.580 & \textbf{0.317} & 7.20 & 7.87 & 8.06 & 7.71 \\
& \persona & \textbf{0.455} & \textbf{0.658} & 0.315 & \textbf{7.35} & \textbf{8.06} & \textbf{8.23} & \textbf{7.88}\\
\addlinespace

\multirow{3}{*}{Gemma3 4B} 
& \beforefinetuning & 0.388 & 0.309 & 0.309 & 6.18 & 6.70 & 7.04 & 6.64  \\
& \syntheticbaseline & 0.408 & 0.510 & 0.323 & 5.29 & 5.93 & 7.34 & 6.22 \\
& \persona & \textbf{0.456} & \textbf{0.643} & \textbf{0.328} & \textbf{6.82} & \textbf{7.65} & \textbf{7.91} & \textbf{7.46} \\
\addlinespace

\multirow{3}{*}{Olmo3 7B} 
& \beforefinetuning & 0.012 & 0.038 & 0.058 & 6.85 & 7.62 & 7.74 & 7.40 \\
& \syntheticbaseline & 0.295 & 0.084 & 0.250 & \textbf{7.08} & 7.96 & 7.99 & 7.67 \\
& \persona & \textbf{0.437} & \textbf{0.590} & \textbf{0.307} & 7.00 & \textbf{7.97} & \textbf{8.16} & \textbf{7.71} \\

\bottomrule
\end{tabularx}
\end{table*}

%% file: tables/test_time_scaling.tex
\begin{table}[!ht]
\centering
\caption{URS Bench results for test-time multi-perspective reasoning.}
\label{tab:urs_only}
\scriptsize
\setlength{\tabcolsep}{3pt}
\begin{tabular}{llcccc}
\toprule
\multirow{2}{*}{\textbf{Model}} & \multirow{2}{*}{\textbf{Variant}} & \multicolumn{4}{c}{\textbf{URS}} \\
\cmidrule(lr){3-6}
& & Leisure & Create & Adv. & \textbf{Avg.} \\
\midrule

\multirow{4}{*}{Qwen3 8B}
& \noprofile       & 6.55 & 7.03 & 6.87 & 6.82 \\
& \syntheticbaseline    & 7.15 & 7.57 & 7.62 & 7.45 \\
& Random   & \textbf{7.37} & \textbf{7.79} & \textbf{7.64} & \textbf{7.60} \\
& \persona    & 7.33 & 7.68 & 7.57 & 7.53 \\
\addlinespace

\multirow{4}{*}{Qwen3 14B}
& \noprofile       & 6.93 & 7.45 & 7.34 & 7.24 \\
& \syntheticbaseline    & 7.23 & 7.70 & 7.56 & 7.49 \\
& Random   & 7.43 & \textbf{7.72} & 7.63 & 7.59 \\
& \persona    & \textbf{7.46} & 7.71 & \textbf{7.65} & \textbf{7.60} \\
\addlinespace

\multirow{4}{*}{Qwen3 32B}
& \noprofile       & 7.05 & 7.17 & 7.00 & 7.07 \\
& \syntheticbaseline    & 7.36 & 7.62 & \textbf{7.56} & 7.51 \\
& Random   & 7.30 & 7.71 & 7.50 & 7.50 \\
& \persona    & \textbf{7.47} & \textbf{7.80} & 7.50 & \textbf{7.59} \\
\addlinespace

\multirow{4}{*}{Gemma3 4B}
& \noprofile       & 6.99 & 7.60 & \textbf{7.66} & 7.42 \\
& \syntheticbaseline    & 7.16 & 7.64 & 7.52 & 7.44 \\
& Random   & 7.24 & \textbf{7.74} & 7.59 & 7.52 \\
& \persona    & \textbf{7.52} & 7.60 & 7.64 & \textbf{7.58} \\
\addlinespace

\multirow{4}{*}{Olmo3 7B}
& \noprofile       & 6.39 & 7.37 & 7.31 & 7.02 \\
& \syntheticbaseline    & 7.22 & 7.60 & 7.57 & 7.46 \\
& Random   & 7.04 & \textbf{7.74} & \textbf{7.69} & 7.49 \\
& \persona    & \textbf{7.28} & 7.62 & 7.66 & \textbf{7.52} \\

\bottomrule
\end{tabular}
\vspace{-0.2in}
\end{table}

%% file: tables/profile_categorical_entropy.tex
\begin{table}[t]
\centering
\small
\setlength{\tabcolsep}{4pt}
\caption{Categorical entropy comparison across three user-profile sets. PRISM serves as baseline. The higher the entropy, the more diverse the distribution.}
\label{tab:categorical-entropy-comparison}
\resizebox{\columnwidth}{!}{%
\begin{tabular}{lccc}
\toprule
Attribute & PRISM & Synthetic & Open-Ended \\
\midrule
Age & 2.462 & 0.644 & 2.129 \\
Gender & 1.116 & 0.516 & 1.196 \\
Employment & 2.279 & 0.319 & 1.870 \\
Education & 2.291 & 0.524 & 2.210 \\
Birth Country & 4.328 & 0.922 & 3.605 \\
Residence Country & 3.766 & 1.156 & 3.638 \\
English Proficiency & 1.484 & 1.004 & 1.963 \\
Marital Status & 1.454 & 0.411 & 1.401 \\
Religion & 1.547 & 0.397 & 0.976 \\
Ethnicity & 1.822 & 1.549 & 1.547 \\
\bottomrule
\end{tabular}%
}
\end{table}

%% file: tables/mauve_score.tex
\begin{table*}[ht!]
\centering
\caption{MAUVE score comparison. Higher score (bolded in the table) indicates closer distributional similarity.}
\label{tab:mauve_input_only}
\small
\begin{tabularx}{\textwidth}{l *{5}{>{\centering\arraybackslash}X}}
\toprule
& & \multicolumn{4}{c}{\textbf{RecBench}} \\
\cmidrule(lr){3-6}
\textbf{MAUVE (0--1)} & \textbf{PRISM} & \textbf{Books} & \textbf{LastFM} & \textbf{Netflix} & \textbf{News} \\
\midrule
Groundtruth vs.\ Open-Ended  & \textbf{0.004704} & \textbf{0.010775} & \textbf{0.004122} & \textbf{0.011518} & \textbf{0.089591} \\
Groundtruth vs.\ Synthetic  & 0.004463          & 0.007296          & 0.004072          & 0.004191          & 0.013032          \\
\bottomrule
\end{tabularx}
\end{table*}

%% file: limitations.tex
\section*{Limitations}
We identify two major limitations of our current study. 
First, we extract our user profiles in a time-agnostic manner, which does not account for the fact that user preference may change over time. As a result, our profiles may miss temporal dynamics that are important for continual deployment in real-world.
Secondly, we extract user profiles solely from Bluesky (due to data license constraints). This means our user profile can be over-representing a certain platform's user group and injects additional bias in downstream tasks.
In the future, we plan to study temporally evolving user profiles and extend our analysis to a more diverse range of data sources.

\section*{Ethical Considerations}
This work utilizes public social media data from the Bluesky platform.
We strictly adhered to the platform's Terms of Service and API use guidelines during the collection of the 2 million posts \cite{dale_2m_bluesky_2024}.
To protect user privacy and mitigate mosaic-effect re-identification, our extraction pipeline explicitly aggregates granular spatiotemporal details into broad behavioral patterns.
No direct quotes or Personally Identifiable Information (PII) are retained in the final user profiles.
The resulting profiles represent generalized behavioral archetypes rather than identifiable human subjects, aligning with standard ethical guidelines for the secondary analysis of public data.

%% file: appendix.tex
\clearpage

\section{PCA Analysis Plots on User Profiles and SFT data}
\label{sec:pca_plots}
\begin{figure}
    \centering
    \includegraphics[width=\linewidth]{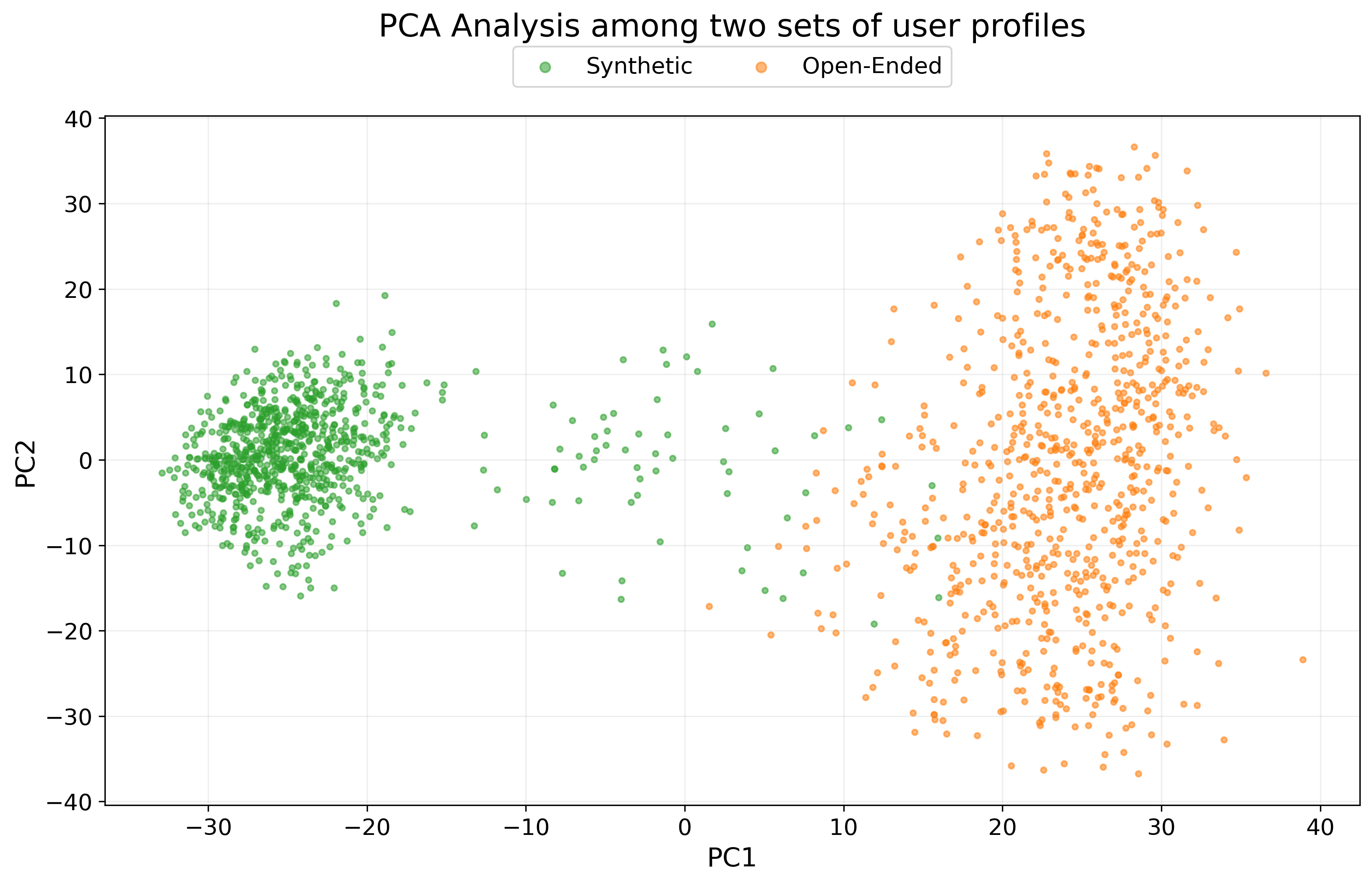}
    \caption{PCA Plot of two sets of user profiles.}
    \label{fig:pca_user_profile}
\end{figure}

\begin{figure*}[p]
    \centering
    \setlength{\tabcolsep}{2pt}

    \begin{subfigure}[t]{0.32\textwidth}
        \centering
        {\scriptsize\textbf{(a)} User profiles\par}
        \vspace{0.2em}
        \includegraphics[width=\linewidth]{figures/pca/user_profiles_two_sets_pca.png}
    \end{subfigure}
    \hfill
    \begin{subfigure}[t]{0.32\textwidth}
        \centering
        {\scriptsize\textbf{(b)} PRISM\par}
        \vspace{0.2em}
        \includegraphics[width=\linewidth]{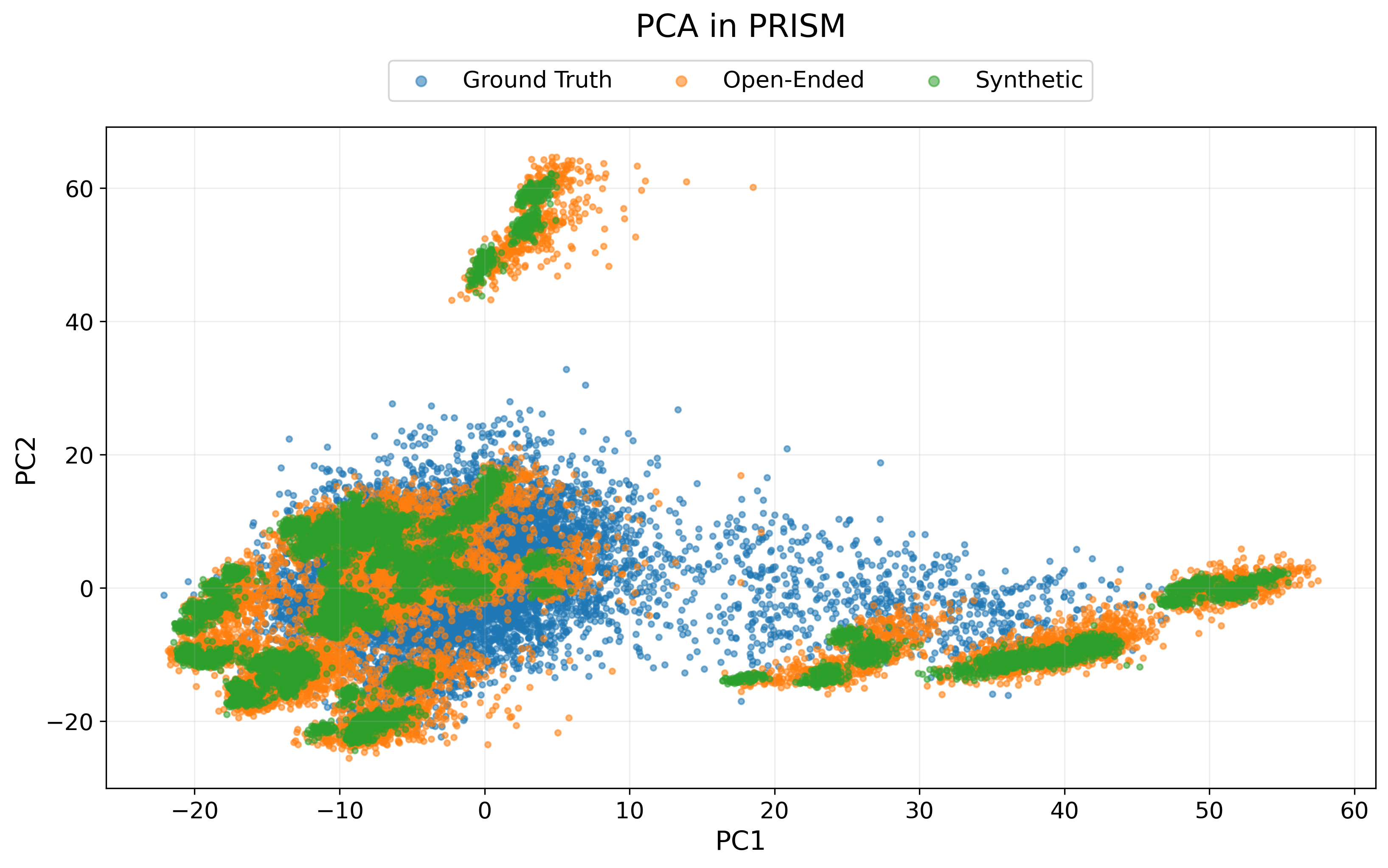}
    \end{subfigure}
    \hfill
    \begin{subfigure}[t]{0.32\textwidth}
        \centering
        {\scriptsize\textbf{(c)} Books\par}
        \vspace{0.2em}
        \includegraphics[width=\linewidth]{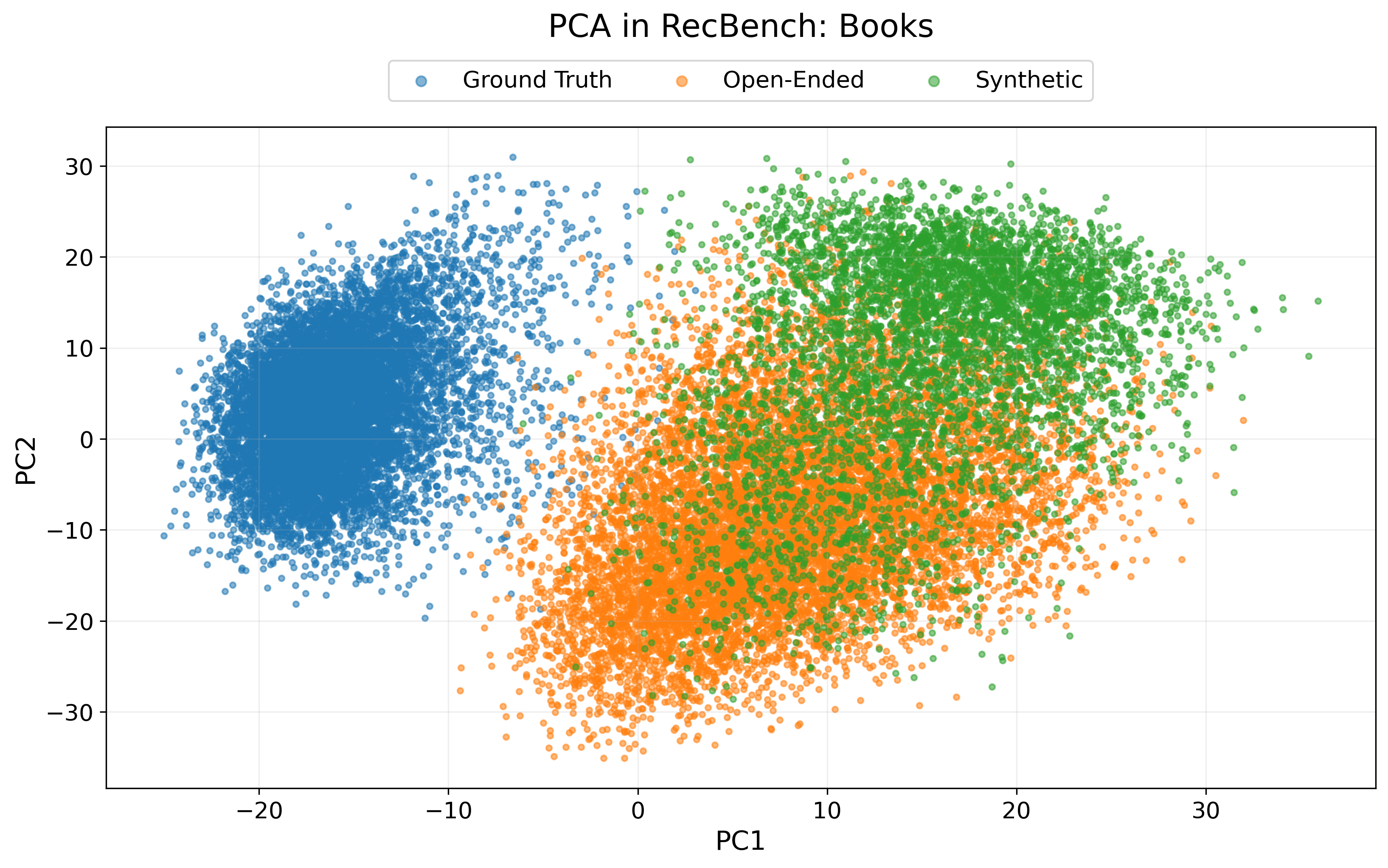}
    \end{subfigure}

    \vspace{0.8em}

    \begin{subfigure}[t]{0.32\textwidth}
        \centering
        {\scriptsize\textbf{(d)} Last.fm\par}
        \vspace{0.2em}
        \includegraphics[width=\linewidth]{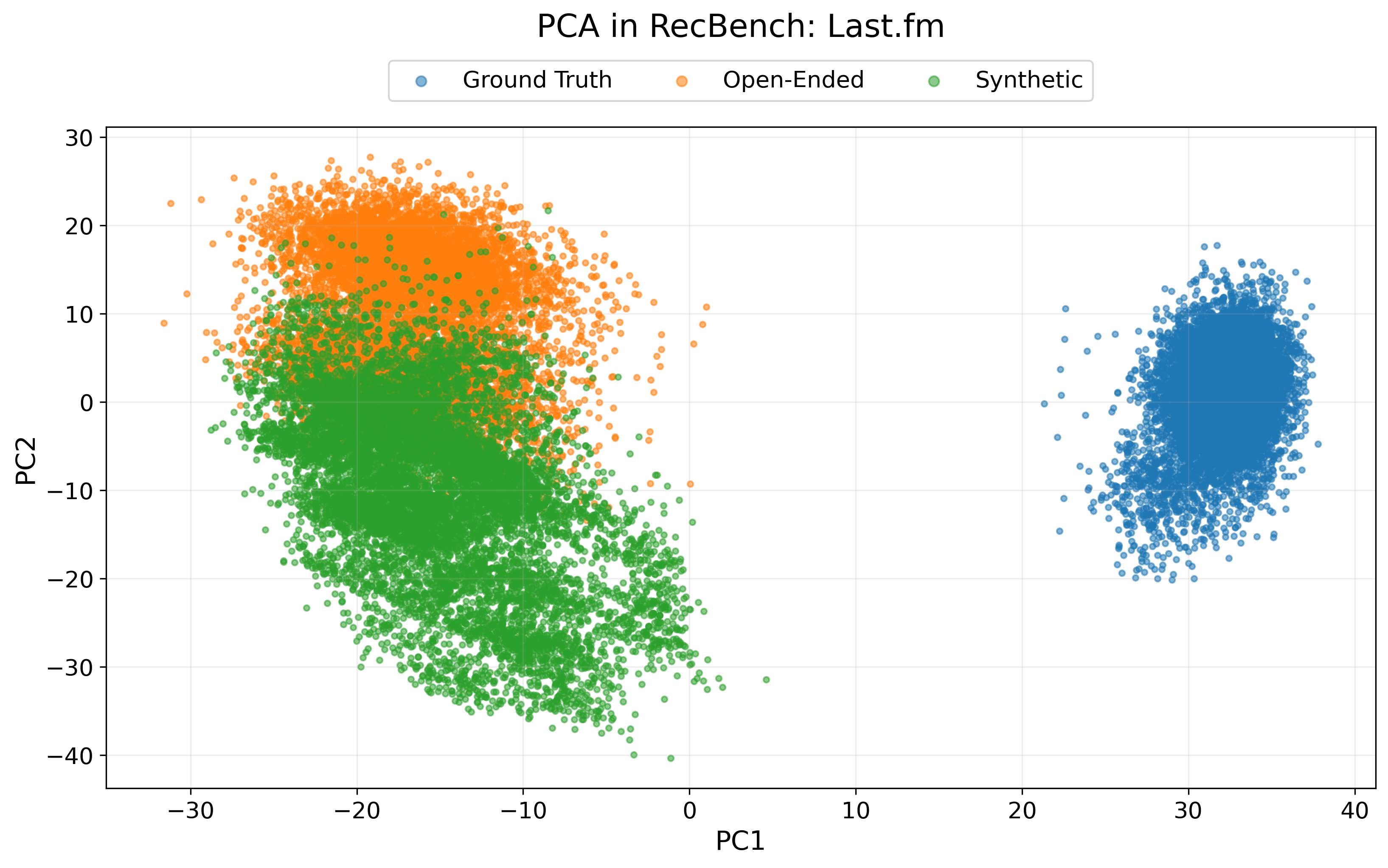}
    \end{subfigure}
    \hfill
    \begin{subfigure}[t]{0.32\textwidth}
        \centering
        {\scriptsize\textbf{(e)} Netflix\par}
        \vspace{0.2em}
        \includegraphics[width=\linewidth]{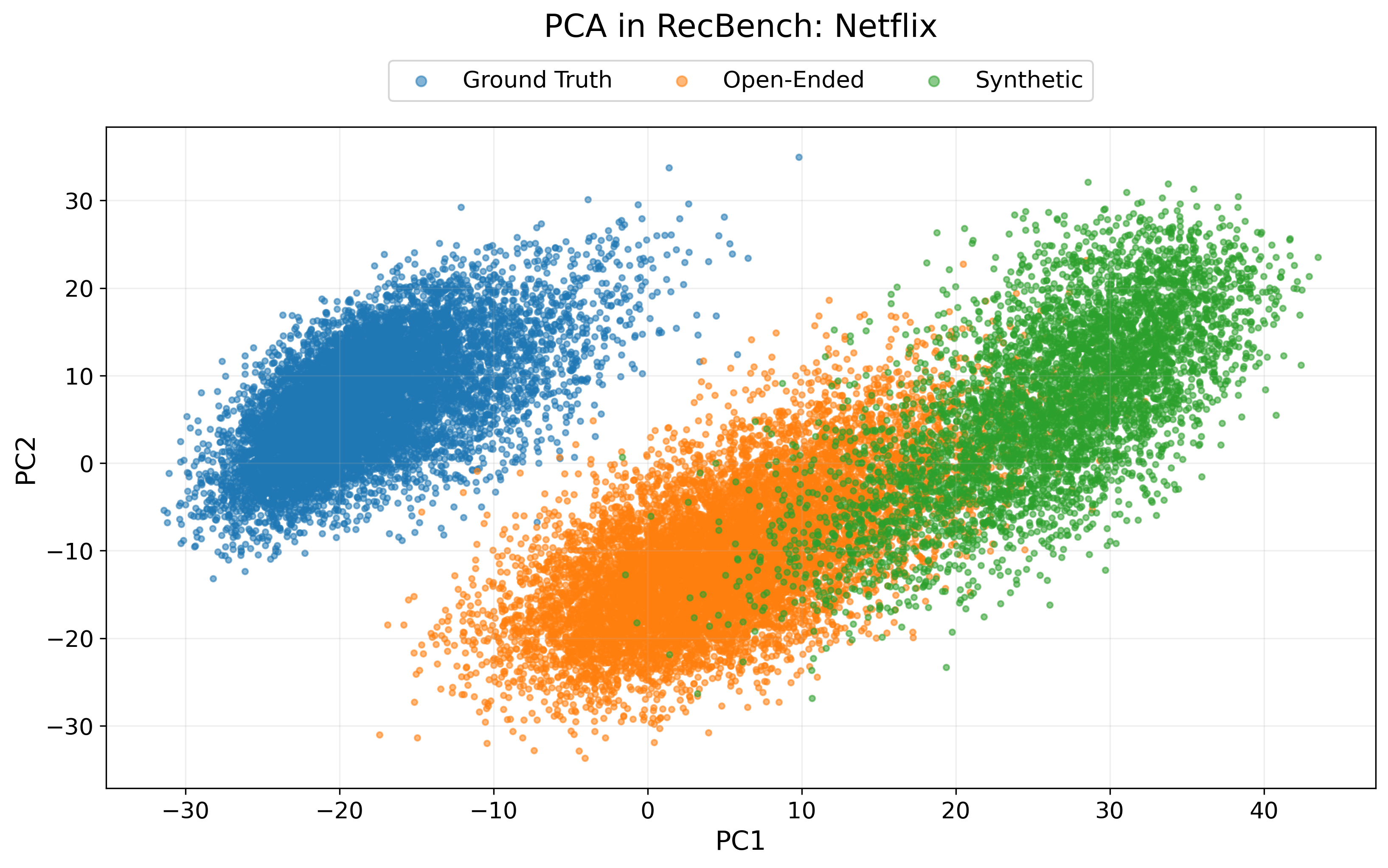}
    \end{subfigure}
    \hfill
    \begin{subfigure}[t]{0.32\textwidth}
        \centering
        {\scriptsize\textbf{(f)} PENS\par}
        \vspace{0.2em}
        \includegraphics[width=\linewidth]{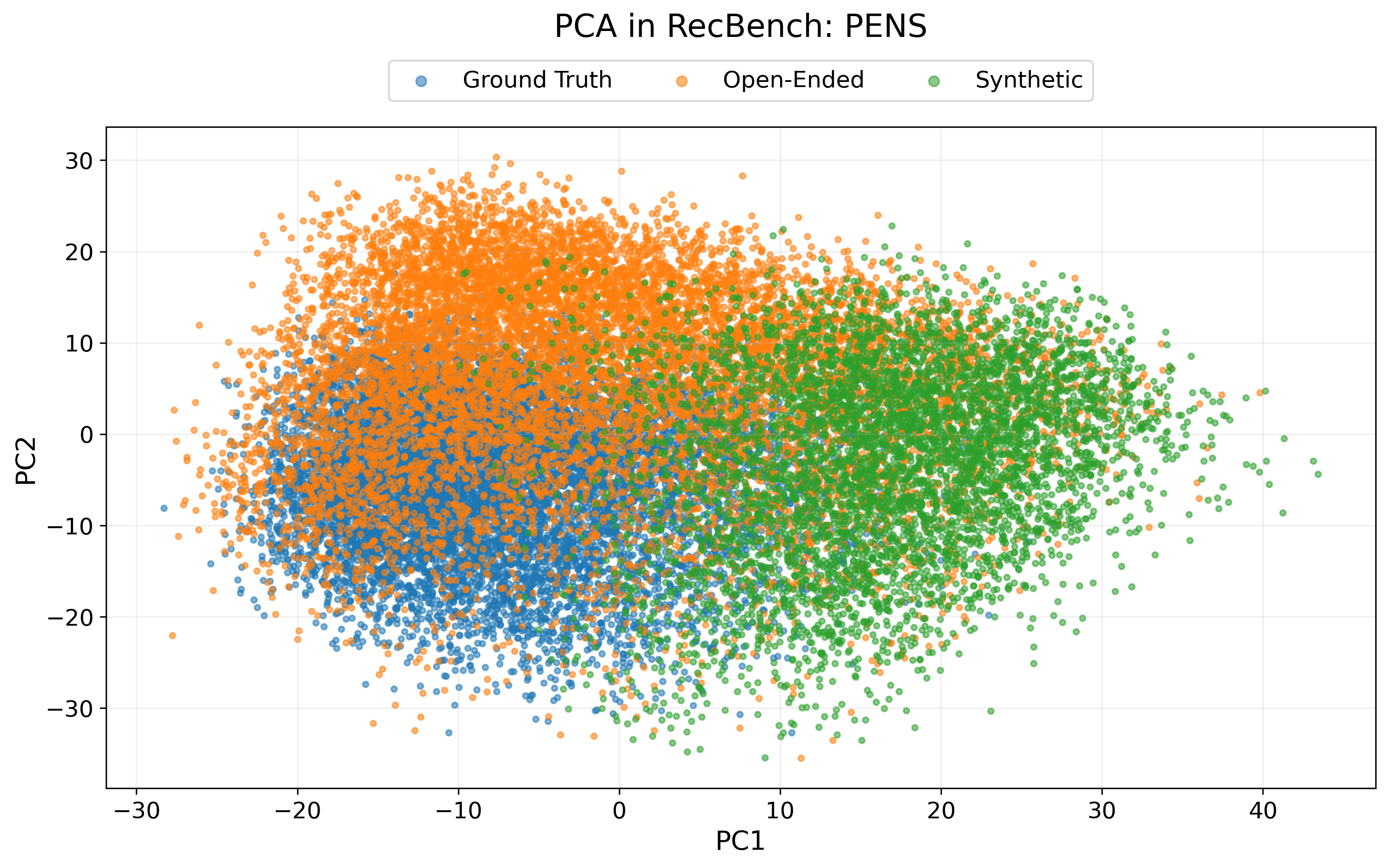}
    \end{subfigure}

    \caption{PCA plots of user-profile and SFT-data embeddings. Panel (a) compares the two sets of user profiles. Panels (b)--(f) show PCA plots of synthesized SFT data embeddings across domains. In almost all domains except Last.fm, open-ended profile-conditioned data exhibits greater diversity while remaining distinguishable from ground-truth data, consistent with the MAUVE results.}
    \label{fig:pca_full_page}
\end{figure*}

The PCA plots of our two sets of user profiles are shown in Figure~\ref{fig:pca_user_profile}. We can see that our user profile (Open-ended) is significantly covering more areas while the synthetic baseline is more homogeneous.
We further present the PCA analysis plots of our synthesized SFT dataset in Figure~\ref{fig:pca_full_page}. In almost all domains (except Lastfm), the Open-ended profile-conditioned data shows greater diversity, while still remaining distinguishable from ground-truth data, consistent with our MAUVE score results.

\section{Experimental Setup}
\label{sec:experimental_details}
Throughout the experiments, we use a total number of 826 user profiles extracted from Bluesky~\citep{dale_2m_bluesky_2024} under Apache license 2.0. For each user profile, we consider up to 100 posts to fit within the LLM's context window.

In our finetuning experiments, we use learning rate of 5e-5, with warmup ratio of 0.1 and weight decay of 0.01. Models are loaded with \texttt{dtype=bfloat16}, and training uses per-device batch size 1 and gradient accumulation 8, for an effective batch size of 8. Each model is finetuned for 1 epoch over 90K SFT examples, with a batch size of 8 on 4 NVIDIA H100 GPUs.
We use full-parameter SFT where it is tractable.
The only exception is Qwen3 32B, where full-parameter tuning was beyond our compute budget, so we employ parameter-efficient SFT using QLoRA~\citep{dettmers2023qloraefficientfinetuningquantized} with rank 64, scaling factor 16, and dropout 0.1 on 8 NVIDIA H100 GPUs.
The Qwen3 32B results follow the same trend as the smaller full-parameter runs; nevertheless, applying QLoRA uniformly across model sizes is a useful robustness check for future work.
For URS Bench, we use 100K examples for SFT training and reserve 92 held-out user profiles for evaluation, ensuring no overlap with the data used for synthesis.

In our test-time scaling experiments, we sample 5 perspectives from 5 user profiles each, due to limited context window space of the aggregator model. We used each model's recommended inference setting. Additionally, we used repetition penalty of 1.15 and frequency penalty of 0.2 to avoid over-lengthy responses.

\section{Negative Results on Lastfm and PRISM}
\label{sec:additional_prism_lastfm}

In this section, we detail two evaluation settings, PRISM preference prediction and Lastfm music recommendation from RecBench, where behaviorally grounded profiles yielded neutral or negative results compared to the baselines.
Analyzing these boundary cases provides valuable insight into the limitations of open-ended profiling, specifically regarding format constraints and domain sparsity.

\paragraph{PRISM Preference Prediction \cite{kirk2024prism}:}
The dataset contains human-annotated preference rankings of LLM conversations, sourced from 1,500 diverse participants who also provided categorical background information. The dataset is under Creative Commons License.
We repurpose this dataset to test whether an LLM can accurately predict a user's preferred response given their profile.
We restrict our evaluation to ``value-guided'' and ``controversy-guided'' conversations, as the ``unguided'' splits are overly generic and rarely require deep personalization. 
In order to match PRISM's categorical user profile, we manually extract attributes such as age, birth location, and gender with an LLM. We do not consider language model usage-series attributes because they have less influence on prediction and are difficult to infer from our open-ended user profiles.
We similarly synthesize SFT data based on our profiles by querying an LLM about whether such a user would prefer a certain model response.
We report the F1 score as the primary metric.

As shown in Table~\ref{tab:task_comparison_lastfm_prism}, we observe mixed outcomes.
While finetuning consistently improves performance over the base models, the behavior-grounded profiles (\persona{}) do not meaningfully outperform the synthetic baselines (\syntheticbaseline{}).
This parity is primarily driven by a structural mismatch between our profiles and the downstream task. Because PRISM natively comes with categorical profiles, we had to perform an additional round of LLM extraction to compress our rich, open-ended bios into a discrete categorical format.
This step inevitably incurred significant information loss, neutralizing the inherent advantage of behavioral grounding.
Consequently, we find that 87.2\% of the synthesized SFT preference pairs were annotated with identical preference labels regardless of whether an open-ended or synthetic profile was used, explaining the underwhelming downstream performance.

\input{tables/sft_appendix_prism_lastfm}

\paragraph{RecBench's Lastfm:}
In the Lastfm music recommendation task, we observe a distinct failure mode, i.e. finetuning degrades the F1 score across the board for both \persona{} and \syntheticbaseline{}.
Our data analysis reveals that this drop stems from three compounding distributional shifts between the synthesized SFT data and the downstream evaluation set:

\begin{itemize}
    \item \textbf{Profile Domain Sparsity:} Up to 30.8\% of our extracted user profiles contain no music-related keywords, and only 8.31\% explicitly mention artist names or track titles. This domain sparsity makes it difficult for the teacher model to synthesize highly relevant music recommendation trajectories.
    \item \textbf{Behavioral Distribution Shift:} Real-world users in the Lastfm evaluation set exhibit strong artist loyalty; 56.4\% of positive candidate items share an artist with the user's interaction history. In contrast, this artist-overlap ratio drops to 24.2\% in the SFT data synthesized from open-ended profiles, and 0\% for the synthetic baseline. Furthermore, the overall distribution of ``popular artists'' hallucinated during SFT diverged significantly from the ground-truth Lastfm distribution.
    \item \textbf{Label Imbalance:} The synthesized SFT data was generated with a balanced class distribution (50\% positive, 50\% negative). However, the downstream Lastfm evaluation set is highly skewed, consisting of 90\% negative examples. Because the finetuned models learned a higher prior for positive interactions, they systematically over-predicted ``Yes'' at inference time, heavily penalizing the final F1 score.
\end{itemize}

\section{How representative are our user profiles?}
\label{app:representative}

To further evaluate the representativeness of our constructed user profiles, we conduct an auxiliary classification study using GPT-OSS-120B. Given a user profile and a post, the model is asked to predict whether the post was written by the user described by that profile. This evaluation tests whether the profile captures enough information to distinguish a user's own posts from those of others. We sampled 200 profiles with 5 posts each, which sum up to 1000 data points in total. All sampled posts are sampled from held-out data and were not used to construct our user profiles.
The model achieves an overall accuracy of 82.80\%, with an accuracy of 83.75\% on negative examples (i.e., posts written by other users). These findings suggest that our user profiles capture salient aspects of the user's post-content distribution.

The prompt we use to conduct this experiment is shown below.
\begin{promptbox}[Prompt for Tweet-Author Matching]
\footnotesize

\textbf{Instruction.}
You are tasked to predict whether a tweet or a series of tweets is from a specific user.
You will be given the user's profile, descriptions, and a list of tweets to evaluate.
Also provide the justification for your answer.
Follow these steps:
(1) summarize the tweet's topic and content, and check whether the topic and content are related to the user's interests in the profile; if not, there is a high chance that the tweet is not from the user;
(2) check whether the tweet's language, writing style, tone, or word usage matches the user's profile; if not, there is a high chance that the tweet is not from the user; and
(3) if multiple tweets are provided, also check their overall pattern, noting that all tweets in the list come from a single user.

\vspace{0.4em}
\textbf{Example}

\textbf{Input}
\begin{quote}
\ttfamily
User Name: <USER NAME>

User Profile: "The user is a Sydney-based Australian business owner working across conveyancing and property valuation (API member) who also runs an Indian/Tandoori catering venture. Client-focused and detail-oriented, they emphasize professionalism, affordability, empathy, and risk management. They follow the Australian property market, central-bank and corporate news, and tax policy (including U.S. property tax). Personal interests include airports/air travel, maritime history, community events and public speaking, and coastal activities around the Northern Beaches; they are spiritually inclined (celebrate Diwali, believe in karma) and occasionally voice support for a local MP."

Tweet 1: Y'all I'm so hype! I'll be performing at my first festival on August 14th!
Be sure to come and check out the performance
Get your tickets today! The link is in my bio
\#luckysummer \#musicfestivals \#independentartist \#livemusic
\end{quote}

\textbf{Output}
\begin{Verbatim}[breaklines=true,breakanywhere=true,fontsize=\scriptsize]
{
  "Justification": "The tweet is about performing at a music festival, which is not related to the user's interests in the profile.",
  "IsFromUser": false
}
\end{Verbatim}

\vspace{0.4em}
\textbf{Task}
Given the following input, predict whether the tweet is from the user.
Only return the results in JSON format.

\begin{quote}
\ttfamily
User Name: <TEMPLATE USER NAME>

User Profile: <TEMPLATE USER PROFILE>

<TEMPLATE TWEETS>
\end{quote}

\end{promptbox}

\section{How faithful are our user profiles?}
\label{app:faithful}
To directly verify that profile content is supported by the underlying behavioral data, we audit profiles at the level of atomic claims. We decompose 50 profiles into 1{,}288 atomic claims and use Claude Opus 4.8---a different model family from our extraction pipeline---to check whether each claim is supported by at least one of the user's original posts. Overall, 88.2\% of atomic claims are grounded in specific posts; a smaller-scale human audit yields a consistent estimate (87.5\% support rate). To rule out hallucination by the LLM auditor itself, we mechanically validate its cited evidence: 93.4\% of verified claim--post links pass a direct substring match against the cited post. These results indicate that the extraction pipeline produces faithful profiles.

\section{Impact of Profile-Conditioned SFT on General Capabilities}
\label{app:general-capabilities}
 
Finetuning on 90K profile-conditioned examples could in principle degrade general-purpose abilities. We therefore evaluate finetuned checkpoints (across RecBench domains) on four standard benchmarks---MMLU (knowledge)~\citep{hendrycks2020measuring}, GSM8K (math)~\citep{cobbe2021training}, IFEval (instruction following)~\citep{zhou2023instruction}, and TruthfulQA (truthfulness)~\citep{lin2022truthfulqa}---for two Qwen3 models, using identical decoding for all checkpoints. Table~\ref{tab:general-capabilities} reports averaged scores and deltas relative to the base models. SFT largely preserves general capability: most deltas are below 4 points, with the worst case at $-5.4$ points (GSM8K, Qwen3-8B). Open-Ended-SFT shifts slightly more than Synthetic-SFT, an expected consequence of stronger specialization, and the trade-off that yields the personalization gains in Table~1 (e.g., $+0.32$ F1 on Books for Qwen3-14B over base). Both regimes remain within the normal range of task-specific SFT.
 
\begin{table*}[h]
\centering
\small
\setlength{\tabcolsep}{3.5pt}
\begin{tabular}{llcccccccc}
\toprule
\textbf{Model} & \textbf{Setting} & \textbf{MMLU} & $\Delta$ & \textbf{GSM8K} & $\Delta$ & \textbf{IFEval} & $\Delta$ & \textbf{TQA} & $\Delta$ \\
\midrule
\multirow{3}{*}{Qwen3-8B}
 & Base           & 0.749 & ---      & 0.876 & ---      & 0.813 & ---      & 0.544 & --- \\
 & Synthetic-SFT  & 0.742 & $-0.007$ & 0.864 & $-0.012$ & 0.828 & $+0.015$ & 0.535 & $-0.009$ \\
 & Open-Ended-SFT & 0.711 & $-0.038$ & 0.823 & $-0.054$ & 0.809 & $-0.004$ & 0.506 & $-0.038$ \\
\midrule
\multirow{3}{*}{Qwen3-14B}
 & Base           & 0.788 & ---      & 0.884 & ---      & 0.850 & ---      & 0.586 & --- \\
 & Synthetic-SFT  & 0.774 & $-0.014$ & 0.860 & $-0.024$ & 0.844 & $-0.006$ & 0.560 & $-0.026$ \\
 & Open-Ended-SFT & 0.765 & $-0.023$ & 0.848 & $-0.036$ & 0.842 & $-0.008$ & 0.542 & $-0.044$ \\
\bottomrule
\end{tabular}
\caption{General-capability benchmarks for base and finetuned models (averaged across RecBench-domain checkpoints), with deltas relative to the base model. SFT largely preserves general capabilities.}
\label{tab:general-capabilities}
\end{table*}

\input{rebuttal_new_experiments}

\input{example_prompts}
\input{example_profiles}

%% file: tables/sft_appendix_prism_lastfm.tex
\begin{table}[!ht]
\centering
\caption{Comparison of F1 scores across base models, models finetuned with task-relevant \syntheticbaseline (Stereotypes), and models finetuned with our open-ended behaviorally grounded profiles. For PRISM, guided corresponds to the guided conversation set. Highest performing scores are highlighted in bold font.}
\label{tab:task_comparison_lastfm_prism}
\small 
\begin{tabular}{llcc}
\toprule
\multirow{2}{*}{\textbf{Model}} & \multirow{2}{*}{\textbf{Variant}} & \textbf{RecBench} & \textbf{PRISM} \\
\cmidrule(lr){3-3} \cmidrule(lr){4-4}
& & LastFM & Guided \\
\midrule

\multirow{3}{*}{Qwen3 8B} 
& \beforefinetuning & \textbf{0.566} & 0.707 \\
& \syntheticbaseline & 0.467 & 0.718 \\
& \persona & 0.358 & \textbf{0.721} \\
\addlinespace 

\multirow{3}{*}{Qwen3 14B} 
& \beforefinetuning & \textbf{0.675} & 0.667 \\
& \syntheticbaseline & 0.165 & \textbf{0.708} \\
& \persona & 0.425 & 0.703 \\
\addlinespace

\multirow{3}{*}{Qwen3 32B} 
& \beforefinetuning & \textbf{0.576} & \textbf{0.701} \\
& \syntheticbaseline & 0.179 & 0.697 \\
& \persona & 0.339 & 0.676 \\
\addlinespace

\multirow{3}{*}{Gemma3 4B} 
& \beforefinetuning & \textbf{0.585} & 0.548 \\
& \syntheticbaseline & 0.165 & \textbf{0.695} \\
& \persona & 0.151 & 0.681 \\
\addlinespace

\multirow{3}{*}{Olmo3 7B} 
& \beforefinetuning & \textbf{0.586} & 0.489 \\
& \syntheticbaseline & 0.515 & \textbf{0.664} \\
& \persona & 0.456 & 0.663 \\

\bottomrule
\end{tabular}
\end{table}

%% file: rebuttal_new_experiments.tex
\clearpage

\section{Reliability Study of the LLM-as-a-judge in URS Bench}
\label{sec:human_judge_validation}
In order to verify the reliability of our LLM, we conduct a human study to verify the agreement between human and LLM judge. We further rescore the responses with an alternative LLM judge to check their consistency.  

\subsection{Human study of LLM Judge}
We conduct a human study to ensure the LLM-as-a-judge evaluation in URS Bench aligns with human judgments of personalization quality.
To this end, we collect 419 pairwise preference judgments from 9 human annotators on gemma3 4b, qwen3 8b model responses (both before and after finetuning). The human annotators are compute science graduate students.
Each annotator compares two model responses to the same URS query under the same profile context and selects the response that is more personalized and useful.
After aggregating human labels with majority vote, the LLM judge agrees with the human-majority preference in 90.14\% of cases, with Cohen's $\kappa=0.8014$.
This high agreement suggests that URS Bench’s LLM-as-a-judge is a reliable proxy for human preferences over personalized responses.

\subsection{Agreement with an Alternative LLM Judge}
To further verify that our conclusions are not tied to GPT-OSS-120B specifically,
we re-score responses with an independent judge from a different model family,
Qwen3-235B-A22B, under the identical rubric and prompts. Over 10k paired judgments,
the Qwen3-235B judge reproduces GPT-OSS-120B's model ranking almost exactly
(Spearman $\rho=0.945$) and correlates strongly per response (Pearson $r=0.81$),
differing mainly by a constant severity offset ($\sim$1 point) that does not affect
relative ordering.

\section{Persona Variations for Personalized Alignment
}
\label{sec:stronger_synthetic_ablation}

We study how the choice of persona representation affects personalized alignment under our profile-conditioned SFT pipeline. 
For this purpose, we first compare \persona{} against stronger synthetic persona sources from prior work. 
Second, we ablate our own profile-construction pipeline by replacing the final profiles with intermediate representations.

\paragraph{Stronger Synthetic Persona Sources}
To test whether our gains depend on a weak synthetic-persona baseline, we compare with two stronger synthetic profile sources: DeepPersona~\citep{wang2025deeppersona} and Nemotron-Personas~\citep{nvidia_nemotron_personas}.

\paragraph{Profile Construction Ablations}
We also evaluate two intermediate representations from our own pipeline: 
\emph{Raw posts} where profiles are formed by simply concatenating a user's original posts, without any pre-processing. 
\emph{Raw descriptors} where we use the extracted behavioral descriptors before deduplication, conflict filtering, and profile summarization.

we run these experiments on RecBench Netflix using Olmo3 7B and Gemma3 4B. 
The results are shown in Table~\ref{tab:profile_representation_comparison}.
\persona{} remains the strongest representation across both evaluated models.
Raw descriptors are close to the final profiles, indicating that the extracted behavioral descriptors already carry much of the useful signal, while the cleaning and summarization stage provides a small but consistent gain and improves readability, conflict control, and prompt cost.

\begin{table*}[t]
\centering
\small
\setlength{\tabcolsep}{6pt}
\resizebox{\textwidth}{!}{
\begin{tabular}{lccccccc}
\toprule
\textbf{Model} & \textbf{\persona{} (Ours)} & \textbf{Raw Descriptors} & \textbf{Raw Posts} & \textbf{DeepPersona} & \textbf{Nemotron} & \textbf{Synthetic} & \textbf{No Profile} \\
\midrule
Olmo3 7B  & \textbf{0.437} & 0.432 & 0.427 & 0.377 & 0.428 & 0.295 & 0.012 \\
Gemma3 4B & \textbf{0.456} & 0.443 & 0.430 & 0.434 & 0.452 & 0.408 & 0.388\\
\bottomrule
\end{tabular}
}
\caption{F1 on RecBench Netflix for stronger synthetic-persona baselines and stage-wise profile ablations.}
\label{tab:profile_representation_comparison}
\end{table*}

\section{Test-Time Multi-Perspective Reasoning: High Temperature Does Not Replace Grounded Perspectives}
\label{sec:temperature_ablation}

We test whether the gains from multi-perspective reasoning require grounded profile-based perspectives.
Specifically, we ask whether they can instead be reproduced by ungrounded output diversity from high-temperature sampling without profile conditioning.
This ablation separates diversity induced by the profile bank from diversity induced by decoding noise.
For this purpose, we sample candidate responses without including any profiles with temperature $T=1.1$ on URS Bench using Qwen3 8B and Olmo3 7B. Then, we compare high-temperature sampling alone (number of answer=1) with aggregated high-temperature candidate set.
As shown in Table~\ref{tab:temperature_ablation}, high-temperature sampling alone performs substantially worse than profile-grounded multi-perspective reasoning.
Aggregation improves over the raw high-temperature samples, but still remains below the profile-grounded scores in Table~\ref{tab:urs_only}.
This indicates that grounded behavioral perspectives provide a stronger source of useful diversity than decoding noise alone.

\begin{table*}[t]
\centering
\footnotesize
\setlength{\tabcolsep}{5pt}
\begin{tabular}{lccccc}
\toprule
\textbf{Model} & \textbf{High-$T$} & \textbf{High-$T$ + Agg.} & \textbf{\persona{} (Ours)} & \textbf{Synthetic} & \textbf{No Profile}\\
\midrule
Qwen3 8B & 5.92 & 7.48 & \textbf{7.53} & 7.45 & 6.82 \\
Olmo3 7B & 6.63 & 7.50 & \textbf{7.52} & 7.46 & 7.02\\
\bottomrule
\end{tabular}
\caption{URS Bench average scores for high-temperature sampling without profile conditioning.}
\label{tab:temperature_ablation}
\end{table*}

%% file: example_prompts.tex
\section{Example Prompts}
\label{sec:prompt}
\subsection{Extracting Behaviorally Grounded User Profiles}
\label{sec:prompt_user_profile}
Below is the prompt example we use to extract short descriptors from the user, which was described in Section~\ref{sec:extraction}. We then further process the acquired information in a short bio.

\begin{promptbox}[Prompt for User Descriptors Extraction from Social Media Posts]
\footnotesize

\textbf{Instruction.}
You are tasked to detect whether the current post contains useful information about the user's background and preferences.
If yes, briefly summarize the possible information revealed about the user.
It is okay to make reasonable assumptions about the user and separate the tags with commas.
If no, or if it is unclear, return ``N/A''.
Also provide the justification for your answer.

\vspace{0.4em}
\textbf{Example 1}

\textbf{Input}
\begin{quote}
\ttfamily
User Name: <USER NAME>

User Post: I took 5,252 photos last night
\end{quote}

\textbf{Output}
\begin{Verbatim}[breaklines=true,breakanywhere=true,fontsize=\scriptsize]
{
  "UsefulInformation": true,
  "Justification": "The user took a lot of photos last night, so they are probably into photography.",
  "UserInformation": "Photography Lover"
}
\end{Verbatim}

\vspace{0.4em}
\textbf{Example 2}

\textbf{Input}
\begin{quote}
\ttfamily
User Name: <USER NAME>

User Post: @<USER NAME> yes
\end{quote}

\textbf{Output}
\begin{Verbatim}[breaklines=true,breakanywhere=true,fontsize=\scriptsize]
{
  "UsefulInformation": false,
  "Justification": "A simple reply to a post with no context",
  "UserInformation": "N/A"
}
\end{Verbatim}

\vspace{0.4em}
\textbf{Example 3}

\textbf{Input}
\begin{quote}
\ttfamily
User Name: <USER NAME>

User Post: "We can reduce risk by regularly washing our hands, not going to work or school if we are sick, and getting the flu shot to reduce overcrowding at health-care facilities during the outbreak." Evergreen advice, really.
\end{quote}

\textbf{Output}
\begin{Verbatim}[breaklines=true,breakanywhere=true,fontsize=\scriptsize]
{
  "UsefulInformation": true,
  "Justification": "The user shares public health advice and emphasizes its importance, suggesting an interest in health, safety, or public awareness.",
  "UserInformation": "Health-Conscious, Values Public Health Advice"
}
\end{Verbatim}

\vspace{0.4em}
\textbf{Task}
Given the following user post, predict whether it contains useful information about the user.
Only return the results in JSON.

\begin{quote}
\ttfamily
User Name: <TEMPLATE USER NAME>

User Post: <TEMPLATE Post>
\end{quote}

\end{promptbox}

Then we would take the information acquired from previous prompt to have a summarized user profile.
\begin{promptbox}[Prompt for User Profile Summarization from User Descriptions]
\footnotesize

\textbf{Instruction.}
You are tasked to summarize a Twitter user's profile given a list of user descriptions.
These descriptions might contain duplicates, vague or unclear statements, or contradictions.
You should follow these steps:
(1) check each description and remove those that do not make sense or are too vague or unclear;
(2) identify similar or duplicate descriptions and merge them into one;
(3) identify descriptions that contradict each other and remove the contradictory ones;
(4) generate a new list of descriptions that is deduplicated and non-contradictory; and
(5) summarize the user profile in a short paragraph describing the user's characteristics and interests based on the cleaned list.

\vspace{0.4em}
\textbf{Example}

\textbf{Input}
\begin{quote}
\ttfamily
User ID: <USER ID>

User descriptions: Australian, Interested in Airports, Possibly Frequent Traveler, Frustrated with Airport Delays...... <More Extracted Contents>
\end{quote}

\textbf{Output}
\begin{Verbatim}[breaklines=true,breakanywhere=true,fontsize=\scriptsize]
{
  "user_id": "<USER ID>",
  "user_description_list": "Australian; Sydney-based; Business account; Business owner/operator (professional services); Conveyancing professional (real-estate law, strata/property transactions); Publishes a conveyancing newsletter; Property valuation/appraisal professional; Australian Property Institute member; Attends industry conferences; Travels for client appointments; Runs Indian/Tandoori catering venture; Offers vegetarian and non-vegetarian options; Provides live catering; Finance and business enthusiast; Follows central bank and corporate news; Interested in tax policy including U.S. property tax; Watches the Australian property market; Collaborates with IT/marketing providers; Uses branding/web and promotional video content; Interested in airports and air travel; Interested in maritime history; Attends events and does public speaking; Client-focused; Values professionalism, affordability, empathy, and risk management; Detail-oriented; Spiritual; Celebrates Diwali; Believes in karma and destiny; Values peace and prosperity; Political supporter of local MP (Mark); Interested in Manly/Northern Beaches; Enjoys coastal/outdoor activities",
  "user_profile": "The user is a Sydney-based Australian business owner working across conveyancing and property valuation (API member) who also runs an Indian/Tandoori catering venture. Client-focused and detail-oriented, they emphasize professionalism, affordability, empathy, and risk management. They follow the Australian property market, central-bank and corporate news, and tax policy (including U.S. property tax). Personal interests include airports/air travel, maritime history, community events and public speaking, and coastal activities around the Northern Beaches; they are spiritually inclined (celebrate Diwali, believe in karma) and occasionally voice support for a local MP."
}
\end{Verbatim}

\vspace{0.4em}
\textbf{Task}
Given the following user descriptions, process them following the same steps and generate the user profile.
Only return the results in JSON.

\begin{quote}
\ttfamily
User ID: <TEMPLATE USER ID>

User descriptions: <TEMPLATE USER DESCRIPTION LIST>
\end{quote}

\end{promptbox}

To serve as a baseline, we construct fake user profiles with the following prompt, which was described in Section~\ref{sec:baselines}.
\begin{promptbox}[Prompt for Fake User Profile Generation]
\footnotesize

\textbf{Instruction.}
You need to generate a fake user profile.
Below are some examples of user profiles.

\vspace{0.4em}
\textbf{Example 1}

\textbf{Output}
\begin{Verbatim}[breaklines=true,breakanywhere=true,fontsize=\scriptsize]
The user is an ......
<Example User Profile>
\end{Verbatim}

\vspace{0.4em}
\textbf{Example 2}

\textbf{Output}
\begin{Verbatim}[breaklines=true,breakanywhere=true,fontsize=\scriptsize]
This user is a .......
<Example User Profile>
\end{Verbatim}

\vspace{0.4em}
\textbf{Task}
Now, generate a fake user profile.
Please ONLY output the generated user profile.

\end{promptbox}

\subsection{Prompt to Synthesize SFT Data}
Below is the prompt we use to annotate potential user interests from user profiles. We then further used the annotation to synthesize SFT data for RecBench finetuning.
\begin{promptbox}[Prompt for Interest Prediction from User Profile]
\footnotesize

\textbf{Instruction.}
You are a recommender.
You are given a <TEMPLATE ITEM TYPE> and a user's profile.
You need to predict whether this user would be interested in this item and provide a justification for your answer.
Please return your answer in JSON format.
If the user is interested in the item, return ``YES''.
If the user is not interested in the item, return ``NO''.
If you are not sure, return ``MAYBE''.

\vspace{0.4em}
\textbf{Example Output Format}

\begin{Verbatim}[breaklines=true,breakanywhere=true,fontsize=\scriptsize]
{
  "Justification": "<your justification for your answer>",
  "IsInterested": "YES"
}
\end{Verbatim}

\vspace{0.4em}
\textbf{Task}
Now, given the following input, predict whether the user would be interested in the item and provide your justification for your answer.
Only output in JSON format.

\begin{quote}
\ttfamily
<TEMPLATE ITEM TYPE>: <TEMPLATE ITEM DESCRIPTION>

User Name: <TEMPLATE USER NAME>

User Profile: <TEMPLATE USER PROFILE>
\end{quote}

\end{promptbox}

We use the following prompt to annotate ``personalized'' answers given user profiles for finetuning in URS Bench, if the user profile is related to the query. Then, profile-conditioned personalized answers are used for SFT in URS Bench.
Note that we use the results for related profiles from the same prompts.

\begin{promptbox}[Prompt for Profile-Grounded Question Answering]
\footnotesize

\textbf{Instruction.}
You are a language model helping a user answer a question.
You are given a user profile and a question.
You need to answer the question based on the user profile.
First, judge whether this user could plausibly ask this question (``MatchUserProfile'').
If not, output ``NO'' and provide the justification.
If yes, output ``YES'' and then answer the question based on the user profile.

\vspace{0.4em}
\textbf{Example 1}

\textbf{Input}
\begin{quote}
\ttfamily
User Profile:
The user is a Southern California resident who enjoys a variety of sweet snacks like chewing gum, novelty candy, biscuits and baked goods. They are a romantic, reflective person who values deep connections and is currently in a relationship. A travel enthusiast, they prefer short trips under two weeks, often planning lavish birthday getaways and traveling with their partner, with a particular interest in Japan. Financially prudent, they focus on savings. Music and club culture matter to them; they love music, appreciate album sequencing, and are nostalgic for the 2009 club scene while being a hip-hop/rap enthusiast. They stay active through fitness, especially Black Sky Fitness, and are a cannabis-using Christian. Active on social media, they use Bluesky, moderate communities, and may subscribe to new platforms. They value realism, continuous learning, history, and literacy, often commenting on social issues. As a senior Black member of AARP, they enjoy big birthday celebrations and milestones."

User Question:

What are the top attractions to visit in Osaka, Japan?
\end{quote}

\textbf{Output}
\begin{Verbatim}[breaklines=true,breakanywhere=true,fontsize=\scriptsize]
{
  "Justification": "The user enjoys travelling and has a particular interest in Japan, so they are likely to ask this question.",
  "MatchUserProfile": "YES",
  "PersonalisedAnswer": "If you're doing a short, romantic Osaka trip with your partner, good options include Dotonbori for food and nightlife, Umeda Sky Building for skyline views, Universal Studios Japan for a big celebration day, Osaka Castle and the Osaka Museum of History for a history-focused visit, Shinsekai for retro city energy, Kaiyukan Aquarium for a calmer date-friendly activity, and Shitenno-ji or Sumiyoshi Taisha for a more reflective stop."
}
\end{Verbatim}

\vspace{0.4em}
\textbf{Example 2}

\textbf{Input}
\begin{quote}
\ttfamily
User Profile:
The user is a German-speaking resident of Germany (likely the Harz region or Munster) with a strong appetite for cultural and culinary traditions, from proverbs to German street food and vegetarian fare. They run a coffee habit, love cooking, and relish specific foods like Harzer cheese, hot dogs, boiled eggs, and tomatoes. Their aesthetic side leans toward timeless design, furniture (especially Vitra), and cozy, comfortable environments. Technologically, they are a power-user who creates AI-generated images, critiques AI chatbots, works as a system administrator/bot developer, and holds logistics and forklift credentials, while staying security-oriented. They are active on social media, track metrics, and inject sarcastic, sometimes vulgar or sexual humor into conversations. Environmentally conscious, they champion sustainability. Politically, they are engaged in German and international law topics, hold conservative, pro-strong-state views, are critical of liberal policies and East German integration, and employ selective content blocking to manage misinformation. They also enjoy cars, value a driver's license, stay health-conscious through walking and outdoor activities, are a parent of young children, and have niche fandoms for Bobo and Turbofreds.

User Question:

Hello, I bought two mattresses from Costco on 12/5/2023, now the mattresses are on sale for \$120 dollars cheaper each. Usually the time to claim the price difference is 30 days, but the courtesy policy I read from reddit is 90 days. What can I tell them in order to get the price difference back?
\end{quote}

\textbf{Output}
\begin{Verbatim}[breaklines=true,breakanywhere=true,fontsize=\scriptsize]
{
  "Justification": "The user is a German resident where Costco does not operate and they use euros instead of dollars, so they are unlikely to ask this question.",
  "MatchUserProfile": "NO",
  "PersonalisedAnswer": "N/A"
}
\end{Verbatim}

\vspace{0.4em}
\textbf{Task}
Now, given the following user profile and question, determine whether the question matches the user profile.
Only return the results in JSON format.

\begin{quote}
\ttfamily
User Profile: <TEMPLATE USER PROFILE>

Question: <TEMPLATE QUESTION>
\end{quote}

\end{promptbox}

\subsection{Prompt for Test Time Scaling}
Below is the prompt to summarise the final answer from sampled perspectives.
\begin{promptbox}[Prompt for Test Time Scaling]
\footnotesize

\textbf{Instruction.}
You are given multiple candidate answers to the same user question.
Write one final answer for the user.
Priority: maximize factual reliability and information coverage while keeping a natural assistant tone.

Hard rules:
(1) Output only a normal user-facing answer, not planning or analysis.
Never write meta text like ``we need to'', ``the task is'', ``across answers'', ``most replies'', ``summary'', or ``integrated''.
(2) Do not add facts not supported by the provided answers.
(3) Prefer details that appear in multiple answers.
If key details conflict, keep the safest shared facts and briefly acknowledge uncertainty instead of guessing.
(4) Remove low-quality content (irrelevant persona assumptions, contradictions, obvious hallucinations, or off-topic refusal text).
(5) Preserve important specifics when supported: names, places, numbers, steps, constraints, caveats, and alternatives.
(6) Follow explicit user constraints exactly (for example, if asked for 5 items, return exactly 5 items).
(7) Never output placeholders or near-empty answers (for example: ``...'', ``N/A'', or blank).
(8) If the question is underspecified (missing a key entity), ask one short clarifying question and still provide the most helpful partial answer possible.
(9) Keep it concise but complete.

\vspace{0.4em}
\textbf{Output Format}
\begin{Verbatim}[breaklines=true,breakanywhere=true,fontsize=\scriptsize]
{
  "summarised_final_answer": "<final user-facing answer>"
}
\end{Verbatim}

\vspace{0.4em}
\textbf{Task}
Given the following question and candidate answers, produce the final answer.
Only return the results in JSON format.

\begin{quote}
\ttfamily
Question: <TEMPLATE QUESTION>

Answers:
<TEMPLATE ANSWERS>
\end{quote}

\end{promptbox}

%% file: example_profiles.tex
\section{Example Profiles and SFT Data}

In this section, we present example synthesized user profiles and downstream supervised fine-tuning (SFT) data instances used in our pipeline.

Below is an example of one of our synthesized user profiles from Bluesky. This is profile is acquired from prompts shown in Appendix~\ref{sec:prompt_user_profile}.

\vspace{0.5em}
\begin{promptbox}[Example of a Synthesized User Profile from Bluesky]
\footnotesize

\textbf{Profile}
\begin{quote}
\itshape
This user is a Canadian veterinarian based in Ottawa with Irish-Canadian roots. They are a mental-health advocate and pet-industry professional who enjoys adventurous activities like caving, diving, and airline experiences. Their interests span navy/maritime deployments, defence news, and a deep engagement with Canadian politics at all levels, including policy areas such as law-enforcement accountability, Indigenous issues, disability rights, and social-justice causes. They follow US and global geopolitics, especially the Middle-East (pro-Israel) and Ukraine-Russia conflicts, and stay informed on economics, inflation, green energy, climate change, sustainability, housing, and the restaurant sector. Humanitarian concerns, natural-disaster updates, and Gaza issues also feature prominently. They monitor journalism, media manipulation, deepfake threats, and cybersecurity, while also curating news and showing interest in IT staffing. Cultural ties include Irish ancestry, Ottawa-Valley heritage, and a love for winter skating.
\end{quote}

\end{promptbox}

\vspace{0.7em}
\begin{promptbox}[Example of a Purely Synthetic User Profile]
\footnotesize

\textbf{Description.}
Below is an example of our purely synthetic user profile. While the content looks no different, the general distribution is much less diverse.

\textbf{Profile}
\begin{quote}
\itshape
The user is a Vancouver-based senior data engineer who splits their weekdays between building scalable ETL pipelines for a fintech startup and mentoring junior developers through a local women-in-tech meetup. Passionate about sustainable tech, they champion carbon-aware cloud architectures and regularly contribute to open-source projects focused on energy-efficient computing. Their digital footprint includes a technical blog where they dissect machine-learning ethics, a curated Twitter feed that mixes data-visualization threads with climate-action commentary, and a monthly newsletter on low-impact coding practices.
\end{quote}

\end{promptbox}

\vspace{0.7em}
\begin{promptbox}[Example of Synthesized SFT Data for RecBench-Books]
\footnotesize

\textbf{Description.}
Below is an example of synthesized SFT data for RecBench Book. The user history is from user profile conditioned annotation. We adopt the same system prompt with Recbench. The answer is YES.

\textbf{Example}
\begin{Verbatim}[breaklines=true,breakanywhere=true,fontsize=\scriptsize]
You are a recommender. I will provide user behavior sequence, and a candidate item. Please response ""YES"" or ""NO"" to represent whether this user is interested in this item. You are not allowed to response any other words for any explanation or note. Now, your role formally begins. Any other information should not disturb you.
(1) Over the Moon A Collection of First Books Goodnight Moon The Runaway Bunny and My World
(2) The Story of Earth The First 45 Billion Years from Stardust to Living Planet
(3) Western Garden Book of Edibles The Complete AZ Guide to Growing Your Own Vegetables Herbs and Fruits
(4) Software Configuration Management Patterns Effective Teamwork Practical Integration
(5) Future Crimes Everything Is Connected Everyone Is Vulnerable and What We Can Do About It
(6) Bowes and Churchs Food Values of Portions Commonly Used
(7) The Missing Piece
(8) Seeds of Earth Humanitys Fire
(9) Confirmation The Hard Evidence of Aliens Among Us
(10) Applewhites at Wits End
(11) Pete the Cat Trick or Pete
(12) Mamas Bank Account HarvestHBJ Book
Candidate item: The Norton Anthology of World Literature Shorter Third Edition  Vol 1
\end{Verbatim}

\end{promptbox}

\vspace{0.7em}
\begin{promptbox}[Example of Annotated Personalized Answer for URS]
\footnotesize

\textbf{Description.}
Below is an example of annotated personalized answer for URS.

\textbf{Input}
\begin{Verbatim}[breaklines=true,breakanywhere=true,fontsize=\scriptsize]
USER PROFILE:
The person is a medical professional--physician, ICU staff member and pulmonary fellow--who is also a medical student and early-career researcher. They advocate for healthcare issues such as ICU visitor policies, physician rights, reproductive rights and public-health policy in Indiana. Alongside their medical work they are a data analyst and visualization enthusiast, highly ad-averse and a farm owner who loves Brussels sprouts. A passionate soccer and football fan, they collect memorabilia and follow international tournaments, while actively fundraising for health-care workers and hunger relief. They stay informed with quality journalism, legal and political news, and champion free-speech and anti-censorship causes. Skeptical of big pharma yet interested in biotech, they promote health-conscious living, nutrition, fitness (plank challenges) and share medical content online. Their personal interests span history, royalty, plant care, fashion-focused TikTok videos, reading (preferring cold settings), video games, 1990s alternative rock, and film set design. Academically driven, they practice time-blocking, productivity techniques and are affiliated with UCSF as a fellow. Based in Indiana (and with ties to Harlem), they are family-oriented, married with occasional relationship uncertainties, and openly discuss depressive and introspective feelings. Known for sarcastic humor, a love of cleanliness, and a current two-week break, they avoid shopping, use Bluesky, and are curious about Twitter verification.

QUESTION:
User Profile: The person is a medical professional--physician, ICU staff member and pulmonary fellow--who is also a medical student and early-career researcher. They advocate for healthcare issues such as ICU visitor policies, physician rights, reproductive rights and public-health policy in Indiana. Alongside their medical work they are a data analyst and visualization enthusiast, highly ad-averse and a farm owner who loves Brussels sprouts. A passionate soccer and football fan, they collect memorabilia and follow international tournaments, while actively fundraising for health-care workers and hunger relief. They stay informed with quality journalism, legal and political news, and champion free-speech and anti-censorship causes. Skeptical of big pharma yet interested in biotech, they promote health-conscious living, nutrition, fitness (plank challenges) and share medical content online. Their personal interests span history, royalty, plant care, fashion-focused TikTok videos, reading (preferring cold settings), video games, 1990s alternative rock, and film set design. Academically driven, they practice time-blocking, productivity techniques and are affiliated with UCSF as a fellow. Based in Indiana (and with ties to Harlem), they are family-oriented, married with occasional relationship uncertainties, and openly discuss depressive and introspective feelings. Known for sarcastic humor, a love of cleanliness, and a current two-week break, they avoid shopping, use Bluesky, and are curious about Twitter verification.
Question: Brainstorm ten names for an orange cat we're adopting from the shelter, with some nickname options as well.
\end{Verbatim}

\textbf{Answer}
\begin{Verbatim}[breaklines=true,breakanywhere=true,fontsize=\scriptsize]
Here are ten orange-cat name ideas that riff on the things you love (medicine, soccer, farms, trivia) -- each comes with a few nickname shortcuts you can use when you're calling the kitty over during those plank-challenge breaks.

 **Cortisol** -- a cheeky nod to your ICU hustle.
   - Nicknames: *Cort*, *Silly*, *Cozy* 

 **Maradona** -- because every orange ball needs a legendary striker.
   - Nicknames: *Mara*, *Donny*, *Dino* 

 **Brussel** -- homage to your favorite sprout-filled garden.
   - Nicknames: *Russ*, *Bee*, *Sprout* 

 **Helix** -- perfect for a data-visualisation nerd who loves twists & turns.
   - Nicknames: *Lex*, *Heli*, *Lixi* 

 **Gryff** -- short for Gryffindor, fitting for a bold feline on a royal-themed TV binge.
   - Nicknames: *Griff*, *Gry*, *FF* 

 **Tango** -- inspired by the rhythm of soccer chants and your love of football.
   - Nicknames: *Tang*, *Go-go*, *Tanny* 

 **Pippin** -- a literary-friendly name (think Tolkien) that sounds cute when you call "Pip!".
   - Nicknames: *Pip*, *Pippy*, *Pin* 

 **Vaxley** -- a playful mash-up of 'vaccine' (big biopharma curiosity) and classic English village vibe.
   - Nicknames: *Vax*, *Ley*, *Val* 

 **Atlas** -- sturdy enough for a farm-owner's companion and evokes world maps you might study in your free-time.
   - Nicknames: *Atty*, *Las*, *Tag* 

 **Pixel** -- ideal for a visualisation aficionado; imagine his fur as tiny orange pixels forming a picture.
   -- Nicknames: *Pix*, *Pexie*, *El*

Feel free to mix-match the nicknames or tweak the base names to suit the kitten's personality once you meet him/her!
\end{Verbatim}

\end{promptbox}

%% file: custom.bib
@article{persona_hub,
  title={Scaling synthetic data creation with 1,000,000,000 personas},
  author={Ge, Tao and Chan, Xin and Wang, Xiaoyang and Yu, Dian and Mi, Haitao and Yu, Dong},
  journal={arXiv preprint arXiv:2406.20094},
  year={2024}
}

@article{prism,
  title={The PRISM alignment dataset: What participatory, representative and individualised human feedback reveals about the subjective and multicultural alignment of large language models},
  author={Kirk, Hannah R and Whitefield, Alexander and R{\"o}ttger, Paul and Bean, Andrew and Margatina, Katerina and Ciro, Juan and Mosquera, Rafael and Bartolo, Max and Williams, Adina and He, He and others},
  journal={Advances in Neural Information Processing Systems},
  volume={37},
  pages={105236--105344},
  year={2024}
}

@article{kirk2024benefits,
  title={The benefits, risks and bounds of personalizing the alignment of large language models to individuals},
  author={Kirk, Hannah Rose and Vidgen, Bertie and R{\"o}ttger, Paul and Hale, Scott A},
  journal={Nature Machine Intelligence},
  volume={6},
  number={4},
  pages={383--392},
  year={2024},
  publisher={Nature Publishing Group UK London}
}

@article{culturellm,
  title={Culturellm: Incorporating cultural differences into large language models},
  author={Li, Cheng and Chen, Mengzhuo and Wang, Jindong and Sitaram, Sunayana and Xie, Xing},
  journal={Advances in Neural Information Processing Systems},
  volume={37},
  pages={84799--84838},
  year={2024}
}

@article{culturepark,
  title={Culturepark: Boosting cross-cultural understanding in large language models},
  author={Li, Cheng and Teney, Damien and Yang, Linyi and Wen, Qingsong and Xie, Xing and Wang, Jindong},
  journal={Advances in Neural Information Processing Systems},
  volume={37},
  pages={65183--65216},
  year={2024}
}

@article{blend,
  title={Blend: A benchmark for llms on everyday knowledge in diverse cultures and languages},
  author={Myung, Junho and Lee, Nayeon and Zhou, Yi and Jin, Jiho and Putri, Rifki A and Antypas, Dimosthenis and Borkakoty, Hsuvas and Kim, Eunsu and Perez-Almendros, Carla and Ayele, Abinew A and others},
  journal={Advances in Neural Information Processing Systems},
  volume={37},
  pages={78104--78146},
  year={2024}
}

@inproceedings{culturalbench,
    title = "{C}ultural{B}ench: A Robust, Diverse and Challenging Benchmark for Measuring {LM}s' Cultural Knowledge Through Human-{AI} Red-Teaming",
    author = "Chiu, Yu Ying  and
      Jiang, Liwei  and
      Lin, Bill Yuchen  and
      Park, Chan Young  and
      Li, Shuyue Stella  and
      Ravi, Sahithya  and
      Bhatia, Mehar  and
      Antoniak, Maria  and
      Tsvetkov, Yulia  and
      Shwartz, Vered  and
      Choi, Yejin",
    editor = "Che, Wanxiang  and
      Nabende, Joyce  and
      Shutova, Ekaterina  and
      Pilehvar, Mohammad Taher",
    booktitle = "Proceedings of the 63rd Annual Meeting of the Association for Computational Linguistics (Volume 1: Long Papers)",
    month = jul,
    year = "2025",
    address = "Vienna, Austria",
    publisher = "Association for Computational Linguistics",
    url = "https://aclanthology.org/2025.acl-long.1247/",
    doi = "10.18653/v1/2025.acl-long.1247",
    pages = "25663--25701",
    ISBN = "979-8-89176-251-0"
}

@inproceedings{casa,
  title={Evaluating cultural and social awareness of llm web agents},
  author={Qiu, Haoyi and Fabbri, Alexander Richard and Agarwal, Divyansh and Huang, Kung-Hsiang and Tan, Sarah and Peng, Nanyun and Wu, Chien-Sheng},
  booktitle={Findings of the Association for Computational Linguistics: NAACL 2025},
  pages={3978--4005},
  year={2025}
}

@article{clo_chat,
  title         = {{CloChat}: Understanding How People Customize, Interact, and Experience Personas in Large Language Models},
  author        = {Ha, Juhye and Jeon, Hyeon and Han, DaEun and Seo, Jinwook and Oh, Changhoon},
  year          = {2024},
  journal       = {arXiv preprint arXiv:2402.15265},
  eprint        = {2402.15265},
  archivePrefix = {arXiv},
  primaryClass  = {cs.HC},
  url           = {https://arxiv.org/abs/2402.15265}
}

@inproceedings{one_chatbot_per_person,
  title={One chatbot per person: Creating personalized chatbots based on implicit user profiles},
  author={Ma, Zhengyi and Dou, Zhicheng and Zhu, Yutao and Zhong, Hanxun and Wen, Ji-Rong},
  booktitle={Proceedings of the 44th international ACM SIGIR conference on research and development in information retrieval},
  pages={555--564},
  year={2021}
}

@inproceedings{personalllm,
  title={PersonaLLM: Investigating the ability of large language models to express personality traits},
  author={Jiang, Hang and Zhang, Xiajie and Cao, Xubo and Breazeal, Cynthia and Roy, Deb and Kabbara, Jad},
  booktitle={Findings of the association for computational linguistics: NAACL 2024},
  pages={3605--3627},
  year={2024}
}

@inproceedings{personalizedllm,
  title={Personalized large language models},
  author={Wo{\'z}niak, Stanis{\l}aw and Koptyra, Bart{\l}omiej and Janz, Arkadiusz and Kazienko, Przemys{\l}aw and Koco{\'n}, Jan},
  booktitle={2024 IEEE International Conference on Data Mining Workshops (ICDMW)},
  pages={511--520},
  year={2024},
  organization={IEEE}
}

@inproceedings{personalizedjudge2024,
  title={Can llm be a personalized judge?},
  author={Dong, Yijiang River and Hu, Tiancheng and Collier, Nigel},
  booktitle={Findings of the Association for Computational Linguistics: EMNLP 2024},
  pages={10126--10141},
  year={2024}
}

@inproceedings{persona_viability,
  title={Generating personas using LLMs and assessing their viability},
  author={Schuller, Andreas and Janssen, Doris and Blumenr{\"o}ther, Julian and Probst, Theresa Maria and Schmidt, Michael and Kumar, Chandan},
  booktitle={Extended abstracts of the CHI conference on human factors in computing systems},
  pages={1--7},
  year={2024}
}

@inproceedings{wang2024selfprompting,
    title = "Unleashing the Emergent Cognitive Synergy in Large Language Models: A Task-Solving Agent through Multi-Persona Self-Collaboration",
    author = "Wang, Zhenhailong  and
      Mao, Shaoguang  and
      Wu, Wenshan  and
      Ge, Tao  and
      Wei, Furu  and
      Ji, Heng",
    editor = "Duh, Kevin  and
      Gomez, Helena  and
      Bethard, Steven",
    booktitle = "Proceedings of the 2024 Conference of the North American Chapter of the Association for Computational Linguistics: Human Language Technologies (Volume 1: Long Papers)",
    month = jun,
    year = "2024",
    address = "Mexico City, Mexico",
    publisher = "Association for Computational Linguistics",
    url = "https://aclanthology.org/2024.naacl-long.15/",
    doi = "10.18653/v1/2024.naacl-long.15",
    pages = "257--279"
}

@inproceedings{wu2024socialsignals,
    title = "Evaluating Large Language Models on Social Signal Sensitivity: An Appraisal Theory Approach",
    author = "Wu, Zhen  and
      Dutt, Ritam  and
      Rose, Carolyn",
    editor = "Soni, Nikita  and
      Flek, Lucie  and
      Sharma, Ashish  and
      Yang, Diyi  and
      Hooker, Sara  and
      Schwartz, H. Andrew",
    booktitle = "Proceedings of the 1st Human-Centered Large Language Modeling Workshop",
    month = aug,
    year = "2024",
    address = "TBD",
    publisher = "ACL",
    url = "https://aclanthology.org/2024.hucllm-1.6/",
    doi = "10.18653/v1/2024.hucllm-1.6",
    pages = "67--80"
}

@inproceedings{patching2025,
  title={Exposing and Patching the Flaws of Large Language Models in Social Character Simulation},
  author={Huang, Yue and Yuan, Zhengqing and Zhou, Yujun and Wang, Xiangqi and Guo, Kehan and Zhuang, Haomin and Sun, Weixiang and Sun, Lichao and Wang, Jindong and Ye, Yanfang and others},
  booktitle={Large Language Models for Scientific and Societal Advances}
}

@article{llmrec,
  title         = {{LLMRec}: Benchmarking Large Language Models on Recommendation Task},
  author        = {Liu, Junling and Liu, Chao and Zhou, Peilin and Ye, Qichen and Chong, Dading and Zhou, Kang and Xie, Yueqi and Cao, Yuwei and Wang, Shoujin and You, Chenyu and Yu, Philip S.},
  year          = {2023},
  journal       = {arXiv preprint arXiv:2308.12241},
  eprint        = {2308.12241},
  archivePrefix = {arXiv},
  primaryClass  = {cs.IR},
  url           = {https://arxiv.org/abs/2308.12241}
}

@article{mgshopdial,
  title         = {{MG-ShopDial}: A Multi-Goal Conversational Dataset for e-Commerce},
  author        = {Bernard, Nolwenn and Balog, Krisztian},
  year          = {2023},
  journal       = {arXiv preprint arXiv:2304.12636},
  eprint        = {2304.12636},
  archivePrefix = {arXiv},
  primaryClass  = {cs.IR},
  url           = {https://arxiv.org/abs/2304.12636}
}

@inproceedings{urs2024,
  title={A user-centric multi-intent benchmark for evaluating large language models},
  author={Wang, Jiayin and Mo, Fengran and Ma, Weizhi and Sun, Peijie and Zhang, Min and Nie, Jian-Yun},
  booktitle={Proceedings of the 2024 Conference on Empirical Methods in Natural Language Processing},
  pages={3588--3612},
  year={2024}
}

@article{liu2025can,
  title={Can LLMs Outshine Conventional Recommenders? A Comparative Evaluation},
  author={Liu, Qijiong and Zhu, Jieming and Fan, Lu and Wang, Kun and Hu, Hengchang and Guo, Wei and Liu, Yong and Wu, Xiao-Ming},
  journal={Advances in Neural Information Processing Systems},
  volume={38},
  year={2025}
}

@article{kirk2024prism,
  title={The prism alignment dataset: What participatory, representative and individualised human feedback reveals about the subjective and multicultural alignment of large language models},
  author={Kirk, Hannah Rose and Whitefield, Alexander and Rottger, Paul and Bean, Andrew M and Margatina, Katerina and Mosquera-Gomez, Rafael and Ciro, Juan and Bartolo, Max and Williams, Adina and He, He and others},
  journal={Advances in Neural Information Processing Systems},
  volume={37},
  pages={105236--105344},
  year={2024}
}

@misc{dale_2m_bluesky_2024,
  author       = {Dale, Alpin},
  title        = {2 Million Bluesky Posts},
  year         = {2024},
  publisher    = {Hugging Face},
  url          = {https://huggingface.co/datasets/alpindale/two-million-bluesky-posts},
  note         = {Dataset repository on Hugging Face. Version: main (commit ba52d3e). Accessed 2026-02-24.}
}

@article{mauvescore,
  title={Mauve: Measuring the gap between neural text and human text using divergence frontiers},
  author={Pillutla, Krishna and Swayamdipta, Swabha and Zellers, Rowan and Thickstun, John and Welleck, Sean and Choi, Yejin and Harchaoui, Zaid},
  journal={Advances in Neural Information Processing Systems},
  volume={34},
  pages={4816--4828},
  year={2021}
}

@inproceedings{ren2025fewshotllmsyntheticdata,
  title={Few-shot LLM synthetic data with distribution matching},
  author={Ren, Jiyuan and Du, Zhaocheng and Wen, Zhihao and Jia, Qinglin and Dai, Sunhao and Wu, Chuhan and Dong, Zhenhua},
  booktitle={Companion Proceedings of the ACM on Web Conference 2025},
  pages={432--441},
  year={2025}
}

@inproceedings{salemi-etal-2024-lamp,
    title = "{L}a{MP}: When Large Language Models Meet Personalization",
    author = "Salemi, Alireza  and
      Mysore, Sheshera  and
      Bendersky, Michael  and
      Zamani, Hamed",
    editor = "Ku, Lun-Wei  and
      Martins, Andre  and
      Srikumar, Vivek",
    booktitle = "Proceedings of the 62nd Annual Meeting of the Association for Computational Linguistics (Volume 1: Long Papers)",
    month = aug,
    year = "2024",
    address = "Bangkok, Thailand",
    publisher = "Association for Computational Linguistics",
    url = "https://aclanthology.org/2024.acl-long.399/",
    doi = "10.18653/v1/2024.acl-long.399",
    pages = "7370--7392"
}

@inproceedings{zhao2025llms,
  title={Do llms recognize your preferences? evaluating personalized preference following in llms},
  author={Zhao, Siyan and Hong, Mingyi and Liu, Yang and Hazarika, Devamanyu and Lin, Kaixiang},
  booktitle={International Conference on Learning Representations},
  volume={2025},
  pages={15888--15931},
  year={2025}
}

@inproceedings{hao2025evaluating,
  title={Evaluating personalized tool-augmented llms from the perspectives of personalization and proactivity},
  author={Hao, Yupu and Cao, Pengfei and Jin, Zhuoran and Liao, Huanxuan and Chen, Yubo and Liu, Kang and Zhao, Jun},
  booktitle={Proceedings of the 63rd Annual Meeting of the Association for Computational Linguistics (Volume 1: Long Papers)},
  pages={21897--21935},
  year={2025}
}

@article{serendipity,
author = {Kotkov, Denis and Wang, Shuaiqiang and Veijalainen, Jari},
title = {A survey of serendipity in recommender systems},
year = {2016},
issue_date = {November 2016},
publisher = {Elsevier},
volume = {111},
number = {C},
doi = {10.1016/j.knosys.2016.08.014},
journal={Knowledge-Based Systems},
month = nov,
pages={180--192},
}

@inproceedings{li2019mind,
  title={Multi-interest network with dynamic routing for recommendation at Tmall},
  author={Li, Chao and Liu, Zhiyuan and Wu, Mengmeng and Xu, Yuchi and Zhao, Huan and Huang, Pipei and Kang, Guoliang and Chen, Qiwei and Li, Wei and Lee, Dik Lun},
  booktitle={Proceedings of the 28th ACM international conference on information and knowledge management},
  pages={2615--2623},
  year={2019}
}

@inproceedings{cen2020controllable,
  title={Controllable multi-interest framework for recommendation},
  author={Cen, Yukuo and Zhang, Jianwei and Zou, Xu and Zhou, Chang and Yang, Hongxia and Tang, Jie},
  booktitle={Proceedings of the 26th ACM SIGKDD international conference on knowledge discovery \& data mining},
  pages={2942--2951},
  year={2020}
}

@inproceedings{cheng2023compost,
  title={CoMPosT: Characterizing and evaluating caricature in LLM simulations},
  author={Cheng, Myra and Piccardi, Tiziano and Yang, Diyi},
  booktitle={Proceedings of the 2023 Conference on Empirical Methods in Natural Language Processing},
  pages={10853--10875},
  year={2023}
}

@misc{dettmers2023qloraefficientfinetuningquantized,
      title={QLoRA: Efficient Finetuning of Quantized LLMs}, 
      author={Tim Dettmers and Artidoro Pagnoni and Ari Holtzman and Luke Zettlemoyer},
      year={2023},
      eprint={2305.14314},
      archivePrefix={arXiv},
      primaryClass={cs.LG},
      url={https://arxiv.org/abs/2305.14314}, 
}

@article{wang2025deeppersona,
  title={Deeppersona: A generative engine for scaling deep synthetic personas},
  author={Wang, Zhen and Zhou, Yufan and Luo, Zhongyan and Ye, Lyumanshan and Wood, Adam and Yao, Man and Mansour, Saab and Pan, Luoshang},
  journal={arXiv preprint arXiv:2511.07338},
  year={2025}
}

@misc{nvidia_nemotron_personas,
  author = {Meyer, Yev and Corneil, Dane},
  title = {{Nemotron-Personas-USA}: Synthetic Personas Aligned to Real-World Distributions
},
  month = {June},
  year = {2025},
  url = {https://huggingface.co/datasets/nvidia/Nemotron-Personas-USA}
  }

@article{richardson2023integrating,
  title         = {Integrating Summarization and Retrieval for Enhanced Personalization via Large Language Models},
  author        = {Richardson, Chris and Zhang, Yao and Gillespie, Kellen and Kar, Sudipta and Singh, Arshdeep and Raeesy, Zeynab and Khan, Omar Zia and Sethy, Abhinav},
  journal       = {arXiv preprint arXiv:2310.20081},
  year          = {2023},
  eprint        = {2310.20081},
  archivePrefix = {arXiv},
  url           = {https://arxiv.org/abs/2310.20081}
}

@article{tan2024democratizing,
  title         = {Democratizing Large Language Models via Personalized Parameter-Efficient Fine-tuning},
  author        = {Tan, Zhaoxuan and Zeng, Qingkai and Tian, Yijun and Liu, Zheyuan and Yin, Bing and Jiang, Meng},
  journal       = {arXiv preprint arXiv:2402.04401},
  year          = {2024},
  eprint        = {2402.04401},
  archivePrefix = {arXiv},
  url           = {https://arxiv.org/abs/2402.04401}
}

@article{zhang2024personalizedmilp,
  title={Personalized llm response generation with parameterized memory injection},
  author={Zhang, Kai and Kim, Yejin and Liu, Xiaozhong},
  journal={arXiv preprint arXiv:2404.03565},
  year={2024}
}

@inproceedings{zhang2025prime,
    title = "{PRIME}: Large Language Model Personalization with Cognitive Dual-Memory and Personalized Thought Process",
    author = "Zhang, Xinliang Frederick  and
      Beauchamp, Nicholas  and
      Wang, Lu",
    editor = "Christodoulopoulos, Christos  and
      Chakraborty, Tanmoy  and
      Rose, Carolyn  and
      Peng, Violet",
    booktitle = "Proceedings of the 2025 Conference on Empirical Methods in Natural Language Processing",
    month = nov,
    year = "2025",
    address = "Suzhou, China",
    publisher = "Association for Computational Linguistics",
    url = "https://aclanthology.org/2025.emnlp-main.1711/",
    doi = "10.18653/v1/2025.emnlp-main.1711",
    pages = "33707--33736",
    ISBN = "979-8-89176-332-6"
}

@article{zhang2026tsubasa,
  title         = {{TSUBASA}: Improving Long-Horizon Personalization via Evolving Memory and Self-Learning with Context Distillation},
  author        = {Zhang, Xinliang Frederick and Wang, Lu},
  journal       = {arXiv preprint arXiv:2604.07894},
  year          = {2026},
  eprint        = {2604.07894},
  archivePrefix = {arXiv},
  url           = {https://arxiv.org/abs/2604.07894}
}

@article{hendrycks2020measuring,
  title={Measuring massive multitask language understanding},
  author={Hendrycks, Dan and Burns, Collin and Basart, Steven and Zou, Andy and Mazeika, Mantas and Song, Dawn and Steinhardt, Jacob},
  journal={arXiv preprint arXiv:2009.03300},
  year={2020}
}

@article{cobbe2021training,
  title={Training verifiers to solve math word problems},
  author={Cobbe, Karl and Kosaraju, Vineet and Bavarian, Mohammad and Chen, Mark and Jun, Heewoo and Kaiser, Lukasz and Plappert, Matthias and Tworek, Jerry and Hilton, Jacob and Nakano, Reiichiro and others},
  journal={arXiv preprint arXiv:2110.14168},
  year={2021}
}

@article{zhou2023instruction,
  title={Instruction-following evaluation for large language models},
  author={Zhou, Jeffrey and Lu, Tianjian and Mishra, Swaroop and Brahma, Siddhartha and Basu, Sujoy and Luan, Yi and Zhou, Denny and Hou, Le},
  journal={arXiv preprint arXiv:2311.07911},
  year={2023}
}

@inproceedings{lin2022truthfulqa,
  title={Truthfulqa: Measuring how models mimic human falsehoods},
  author={Lin, Stephanie and Hilton, Jacob and Evans, Owain},
  booktitle={Proceedings of the 60th annual meeting of the association for computational linguistics (volume 1: long papers)},
  pages={3214--3252},
  year={2022}
}
